\documentclass[journal]{IEEEtran}
\IEEEoverridecommandlockouts
\usepackage{cite}
\usepackage{amsmath,amssymb,amsfonts}
\usepackage{minipage-marginpar}
\usepackage[font=small]{caption}
\usepackage{enumitem} 
\usepackage{subfig}        
\usepackage{graphicx}
\usepackage{textcomp}
\usepackage{xcolor}
\usepackage{hyperref}       
\usepackage{url}            
\usepackage{booktabs}       
\usepackage{amsfonts}       
\usepackage{mathtools}
\usepackage{float}
\usepackage{dblfloatfix}    
\usepackage{multirow}
\usepackage{algpseudocode}
\usepackage[ruled,vlined,linesnumbered]{algorithm2e}

\usepackage{amsthm}
\usepackage{etoolbox}

\AfterEndEnvironment{theorem}{\noindent\ignorespaces}

\AfterEndEnvironment{corollary}{\noindent\ignorespaces}

\AfterEndEnvironment{lemma}{\noindent\ignorespaces}
\newtheorem{definition}{\textbf{Definition}}
\AfterEndEnvironment{definition}{\noindent\ignorespaces}

\AfterEndEnvironment{example}{\noindent\ignorespaces}
\usepackage{amssymb}
\usepackage{amsmath}

\usepackage{xfrac}
\usepackage{array}
\newcolumntype{x}[1]{>{\centering\arraybackslash\hspace{0pt}}p{#1}}

\def\BibTeX{{\rm B\kern-.05em{\sc i\kern-.025em b}\kern-.08em
		T\kern-.1667em\lower.7ex\hbox{E}\kern-.125emX}}
\begin{document}

%
\title{COVID-19 and Your Smartphone: BLE-based Smart Contact Tracing}

\author{Pai Chet Ng,~\IEEEmembership{Student Member,~IEEE,}
    Petros Spachos,~\IEEEmembership{Senior Member,~IEEE,} and \\
    Konstantinos N. Plataniotis,~\IEEEmembership{Fellow,~IEEE}
    \thanks{Pai Chet Ng is with the Department of Electronics and Computer Engineering, Hong Kong University of Science and Technology, Hong Kong. E-mail: pcng@ust.hk}
    \thanks{Petros Spachos is with the School of Engineering, University of Guelph, Canada. E-mail: petros@uoguelph.ca}%
    \thanks{Konstantinos N. Plataniotis is with the Department of Electrical and Computer Engineering, University of Toronto, Canada. E-mail: kostas@ece.utoronto.ca}
}



\maketitle
\begin{abstract}
Contact tracing is of paramount importance when it comes to preventing the spreading of infectious diseases. Contact tracing is usually performed manually by authorized personnel. Manual contact tracing is an inefficient, error-prone, time-consuming process of limited utility to the population at large as those in close contact with infected individuals are informed hours, if not days, later. This paper introduces an alternative way to manual contact tracing. The proposed Smart Contact Tracing (SCT) system utilizes the smartphone’s Bluetooth Low Energy (BLE) signals and machine learning classifier to accurately and quickly determined the contact profile. SCT’s contribution is two-fold: a) classification of the user’s contact as high/low-risk using precise proximity sensing, and b) user anonymity using a privacy-preserving communications protocol.  SCT leverages BLE’s non-connectable advertising feature to broadcast a signature packet when the user is in the public space. Both broadcasted and observed signatures are stored in the user’s smartphone and they are only uploaded to a secure signature database when a user is confirmed by public health authorities to be infected.  Using received signal strength (RSS) each smartphone estimates  its distance from other user’s phones and issues real-time alerts when social distancing rules are violated.  The paper includes extensive experimentation utilizing real-life smartphone positions and a comparative evaluation of five machine learning classifiers.  Reported results indicate that a decision tree classifier outperforms other states of the art classification methods in terms of accuracy. Lastly, to facilitate research in this area, and to contribute to the timely development of advanced solutions the entire data set of six experiments with about 123,000 data points is made publicly available. 
\end{abstract}


%
\IEEEpeerreviewmaketitle

\section{Introduction} 
\label{sec:intro}
\noindent
\IEEEPARstart{C}{ontact} tracing is an important step in containing a disease outbreak~\cite{8587238},~\cite{eames2003contact}. Many efforts have been devoted to tracing a list of contacts when a person is diagnosed with a highly infectious disease, such as COVID-19. The current contact tracing method, which requires a collaborative effort from several authorized personnel, is labor-intensive and time-consuming~\cite{mitReviewCT}. Since it takes time to trace the contact, the group of users who have been in contact with an infected individual might spread the disease to another group of people before they get informed. It is critical to have an effective contact tracing method that not only can automatically inform the potential users immediately but also reducing the required amount of labor force~\cite{ferretti2020quantifying}.

\begin{figure}[t!]
	\centering
	\includegraphics[width=0.55\columnwidth]{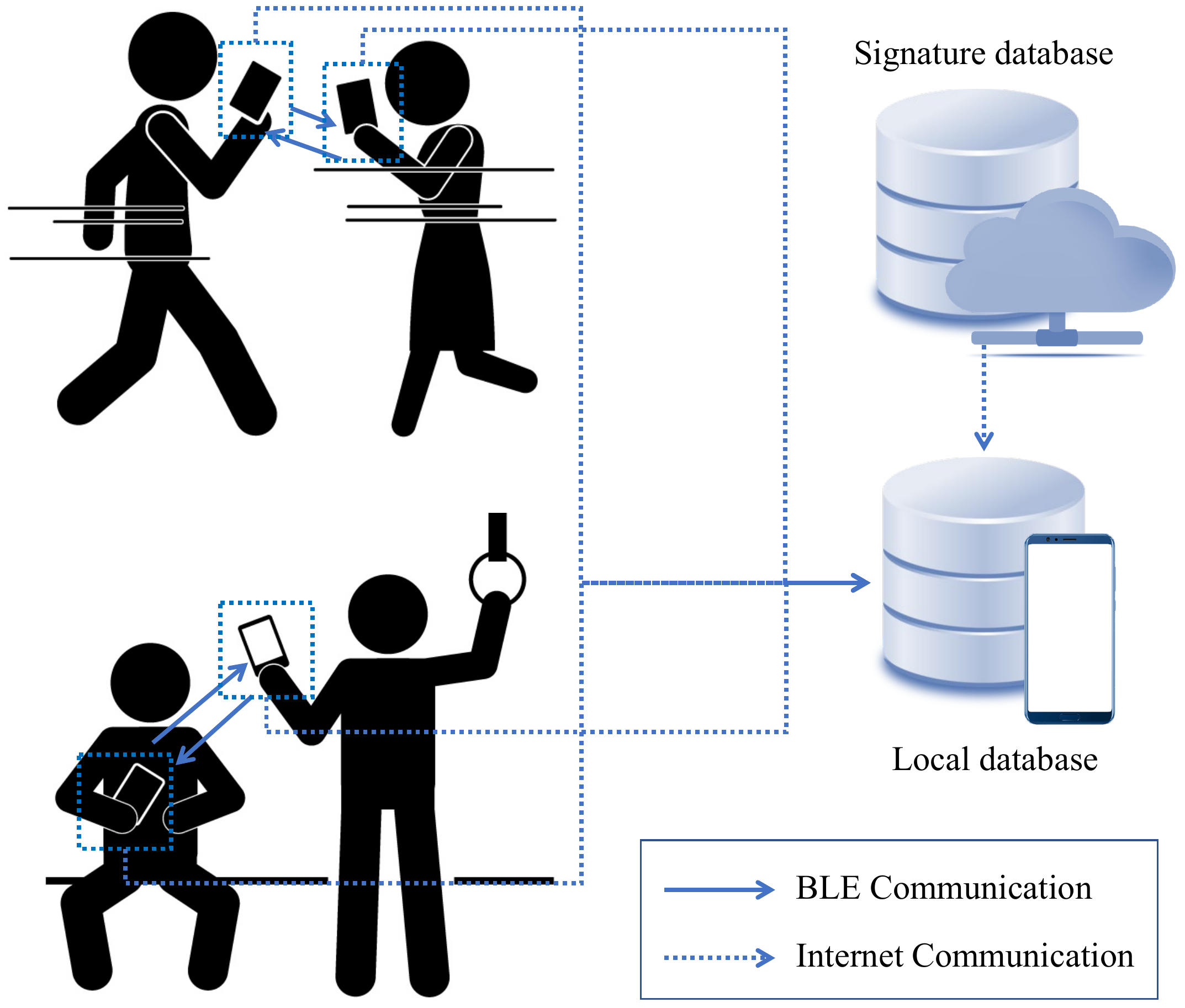}
	\caption{Smart Contact Tracing (SCT) system that uses non-connectable BLE transmission to periodically broadcast a constantly-changing signature packet.} 
	\label{fig:theSystem}
\end{figure}

To this end,  a smart contact tracing (SCT) system is introduced by exploiting the Bluetooth Low Energy (BLE) signals on smartphones.
BLE is ubiquitous and is readily available on many smartphones making it ideal for the introduced system~\cite{7366936}. On the other hand, smartphones have become an intimate device in our everyday life. While we might leave the smartphone  away from us when we are in our private space (e.g., home, private office, etc.), we always carry the smartphone when we do the grocery shopping, commute on public transport, walk along the open street, etc. In this way, smartphones are the best choice for contact tracing, in which the tracing is only performed when the user is in the public space. An overall illustration of our introduced SCT system is shown in Fig.~\ref{fig:theSystem}.   At any time, no location or any other information regarding the users is collected or transmitted. The application uses only BLE signals and no information exchange. The system has three main objectives: preserve-privacy, provide accurate contact tracing, and provide real-time proximity alerts.

\textit{Preserve- privacy.} We leverage the beaconing feature in BLE wireless technology to broadcast an encrypted packet periodically~\cite{8242361}. This encrypted packet is broadcast on the non-connectable advertising channels (i.e., CH 37, 38, and 39). Hence, our proposed SCT system can prevent unauthorized access to the user's smartphone. Furthermore, the packet encrypts a piece of unique signature information based  on the ambient environmental features the smartphone encountered at a particular time. This signature is unique and is almost impossible to be duplicated by another device on another occasion.
All the broadcast signatures and observed signatures will be stored in the local storage.
The user is only required to upload their own broadcast signatures to the signature database when the user is confirmed to be infected with the contagious disease.
Otherwise, the signature store in the local storage will be deleted automatically when it is expired.
We define the signature expiration according to the disease spreading time window, as suggested by the health authorities.
By comparing the signature of each smartphone, a list of possible contacts can be retrieved without explicitly revealing the sensitive information of the infected user.

\textit{Accurate contact tracing.} The smartphone application identifies contacts in proximity, over time.  It records the estimated distance and the duration of interaction between individuals. In this way, it will identify when someone has been too close to an infected person for too long (the too close for too long (TC4TL) problem). For instance, when people hug each other they are too close for a  short period of time, while inside the cabin of a flight, people can be in ten meters distance for too long, breathing the same air. At the same time, a distance of two meters in a classroom might be safe while three meters in a subway train might trigger an alert. We have to rely heavily on the virologists and the epidemiologists to identify the healthy distance in different environments, and through this application, we can give them access to these crucial details. 

\textit{Real-time proximity alerts.} The application will provide a real-time alert to the user if the physical distancing between any two users is not maintained. This will be achieved by detecting the proximity between any users in a given location, including the grocery store, public transit, etc. This proximity information can be retrieved by inspecting the RSS patterns from the user's smartphone~\cite{8423010}. A smartphone can measure the RSS value upon seeing the packet broadcast by the nearby smartphone. Since RSS is inversely proportional to the square of the distance~\cite{8395148}, we can use it to estimate the distance between any two smartphones and then classify the proximity based on the recommended physical distancing rule. 

Precise distance estimation through the RSS is necessary to determine the proximity between any two smartphones.  However, RSS is subject to severe fluctuation especially the body shadowing effect since the smartphone is carried by users~\cite{6856188}. We examine five machine learning-based classifiers: decision tree (DT), linear discriminant analysis (LDA), naive bayes (NB), $k$ nearest neighbors (kNN), and support vector machine (SVM), over six different smartphone positions: Hand-to-Hand (HH), Hand-to-Pocket (HP), Hand-to-Backpack (HB), Pocket-to-Backpack (PB), Pocket-to-Pocket (PP), and Backpack-to-Backpack (BB).

In summary, this paper has the following contributions:
\begin{itemize} 
    \item Privacy-preserving signature protocol: our SCT system provides a secure contact tracing by using the non-connectable advertising channels and an encrypted packet containing unique signature information based on the ambient environmental features observed by a smartphone.
    \item Proximity sensing  and real-time physical distance alert, with precise distance estimation: we classify the proximity of a user to any user by estimating the distance between any two users based on the RSS values measured by each smartphone, while we push a notification to alert the users when anyone violates the physical distancing rule. After approximately 10~s of interaction between smartphones, the system is able to provide a reliable estimation. DL is the most accurate classifier.
    \item Smartphone implementation and effects of smartphone's position: we prototyped our system design and implemented the application in modern smartphones to demonstrate the feasibility of our proposed SCT. The energy requirements of the application are negligible. We compared the classifiers in terms of their estimation accuracy, while we examine six different positioning sets of smartphones. When the users have their smartphones in similar positions, the classifiers can improve accuracy.
    \item Extensive experiments: we performed extensive experiments in a real-world setting verify the effectiveness of our SCT. All the collected data is available in the IEEE Dataport~\cite{ieee-dataportRSS} and Github~\cite{githubRSS}. The overall dataset contains the measurement data obtained from six experimental sets, amounting to a total of 123,718 data points. We believe that the dataset will serve as an invaluable resource for researchers in this field, accelerating the development of contact tracing applications.
\end{itemize}


\section{Background, Motivation and Components}
\label{sec:background}
Contact tracing aims to track down a group of users who have encountered with an infected individual.
The goal is to inform this group of users regarding the potential risk that they might face so that they can take appropriate actions as recommended by the local health authority.

\subsection{Current Development in Contact Tracing}

\begin{figure}[t!]
\centering
\subfloat[]
{\includegraphics[width=.6\columnwidth]{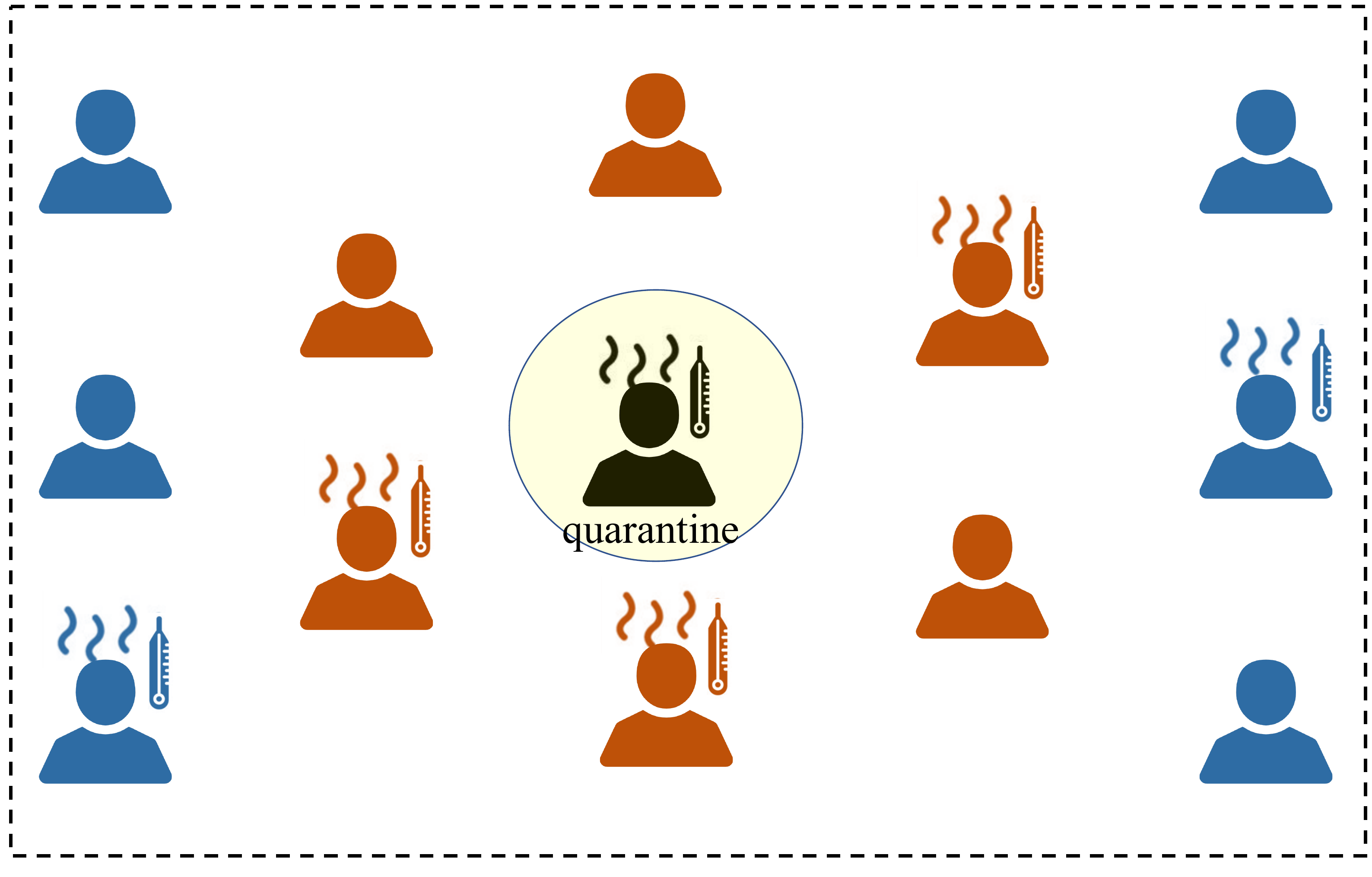}\label{fig:s1_f}}

\subfloat[]
{\includegraphics[width=.6\columnwidth]{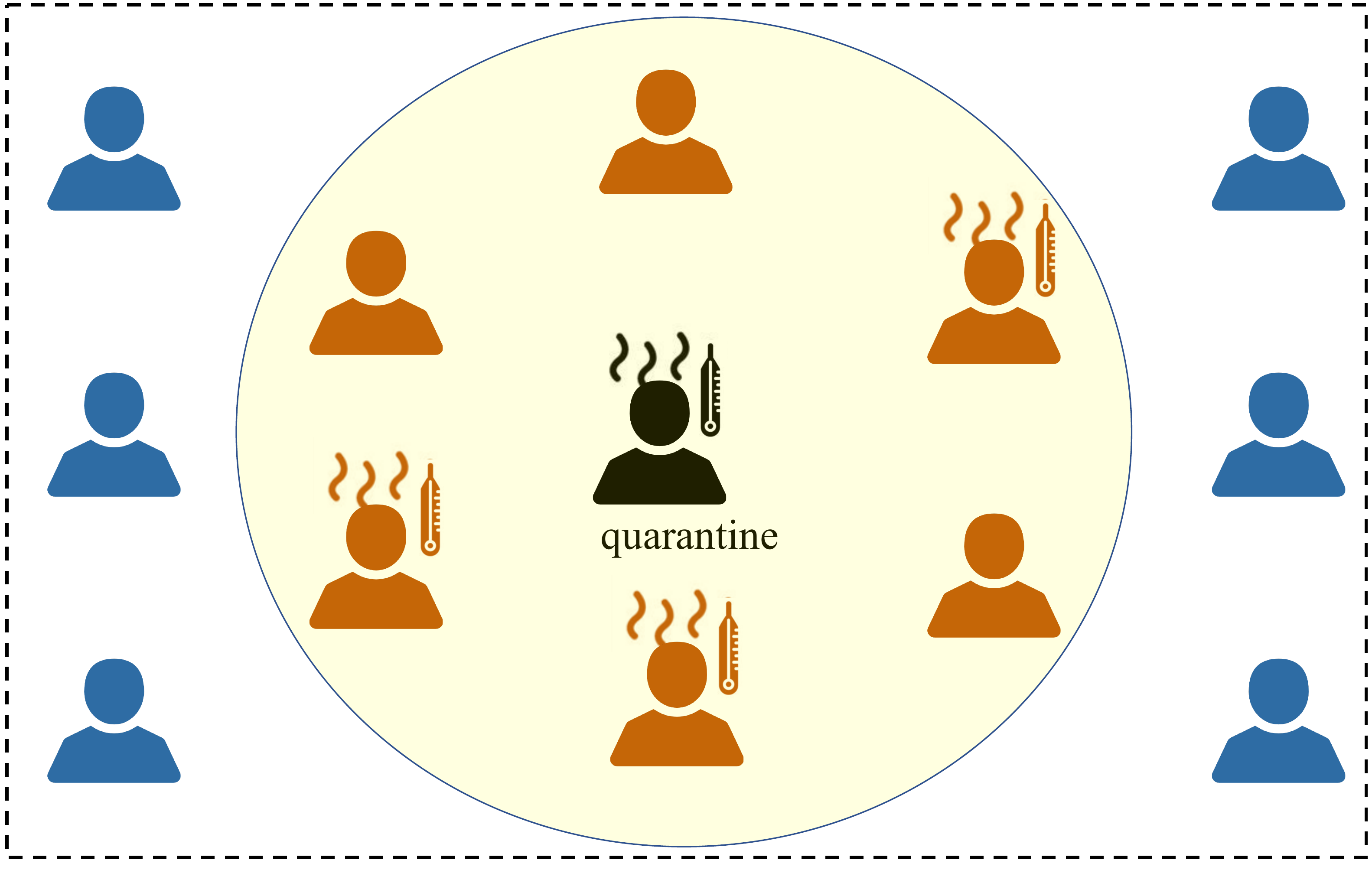}\label{fig:s1_t}}
\caption{ Spread of an infectious disease: (a) only quarantine the infected individual, (b) quarantine the infected individual and all the users who have been in contact with the infected individual.}
\label{fig:contactTracing}
\vspace*{-0.5cm}
\end{figure}

Contact tracing could be a viable solution in resuming the normal lifestyle while preventing the further virus outbreak. An illustration of the differences in having a contact tracing system is shown in Fig.~\ref{fig:contactTracing}. In practice, a person will be quarantined immediately when they are confirmed to be infected with the disease. However, those people who have been in close contact with the infected individual are still free to move without realizing that they may have already got infected and became a virus carrier. With contact tracing, we can inform most of the potential close contact so that they can take appropriate action to isolate themselves from the crowd.

Recognizing the importance of contact tracing, many countries have put effort to develop a smartphone-based contact tracing system.
\begin{itemize}
	\item China: In China, a close contact detector based on QR codes technology is implemented~\cite{chinaCorona}. 
	The  application is developed based on a surveillance strategy in monitoring the people's movement within the country and it can push an alert to users if they have been in close contact with the infected individual.
	\item South Korea: In South Korea, the location data (i.e., the GPS data) obtained from the user's smartphone are used to detect the distance of the users from the infected individual. The tracker application will push a notification that contains the personal details of the infected individuals to the potential users who have been in contact with the infected individual~\cite{skCorona}.
	\item Singapore: In Singapore, a privacy-preserving approach is adopted, by using the BLE signal on the smartphone to detect the proximity between any two individuals. The TraceTogether application broadcasts an encrypted packet, which is generated by a secret key distributed by the Ministry of Health, given the phone number~\cite{sgCorona}. The application will also alert the users when they are in close contact with the infected individual.
\end{itemize}

The first two approaches might  compromise user's privacy since their applications require to monitor the user's mobility and locations. On the other hand, in the third approach, although it preserves privacy by tracking only the proximity between users without explicit location information, the encryption process involves the user's phone number. Hence, the phone number might be retrieved by a malicious hacker.

Besides the national-level effort, there are collaborations in industry and academia  in delivering an effective contact tracing solution while preserving user privacy~\cite{shukla2020privacy},~\cite{bell2020tracesecure}. Rather than using location data, many of these initiatives focus on the use of BLE signals for proximity detection. For instance, Pan European Privacy-Preserving Proximity Tracing (PEPP-PT) detects the proximity based on the broadcast BLE packet containing a full anonymous ID~\cite{PEPP}. COVID-19 Watch, on the other hand, can automatically alert the user when they are in contact with the infected individual~\cite{COVIDwatch}. Similarly, the Privacy-Preserving Automated Contact Tracing (PACT) exploits the BLE signals in combination with secure encryption to detect possible contacts while protecting privacy~\cite{PACT}.

Most of the above initiatives assume that the BLE signals will work for proximity detection without considering the smartphone positioning effects on it.  To the best of our knowledge, there is a lack of a comprehensive study on the accuracy of BLE signals for contact tracing proximity sensing. Furthermore, most of the encryption methods are based on information provided by the user, which might be subject to possible information leaks if the encryption method is compromised. To bridge the gap, this paper studies the proximity sensing with the BLE signals broadcast from the smartphones carried by the user while designing a privacy-preserving signature protocol that uses the environmental feature instead of user information for packet broadcasting. Six experimental sets, with different smartphone positioning are examined, to investigate the feasibility of the system under different realistic conditions.

\subsection{Bluetooth Low Energy (BLE) Technology}
\label{ss:ble}
BLE provides a short-range communication over the 2.4~GHz ISM band~\cite{gomez2012overview}. It is ubiquitous and has been adopted by many smart devices (e.g., smartwatches, earphones, smart thermostats, etc.) as the main communication platform~\cite{7000963}. Furthermore, BLE is readily available in most modern smartphones regardless of the operating system. There are two modes of communication available with BLE: 1)~non-connectable advertising, and 2)~connectable advertising~\cite{8011489}.
The latter advertising mode allows another device to request a connection by sending a CONNECT\_REQ packet on the advertising channels.
In this work, we focus on the non-connectable advertising mode, in which the device cannot accept any incoming connection requests. This feature is useful for our SCT system in ensuring no neighboring devices can access the smartphone to retrieve sensitive information. 
For contact tracing purposes, we configure the smartphone to periodically broadcast the advertising packet via the non-connectable advertising mode. These packets can be heard and received by any nearby smartphones as long as these smartphones are within the broadcast range. These smartphones can also measure the received signal strength (RSS) upon receiving the packet. 

However, there are two major challenges: 1)~the length of the advertising packet is only up to 47~bytes, and 2)~the RSS values are subject to severe fluctuation.

\begin{figure}[t!]
	\centering
	\includegraphics[width=0.85\columnwidth]{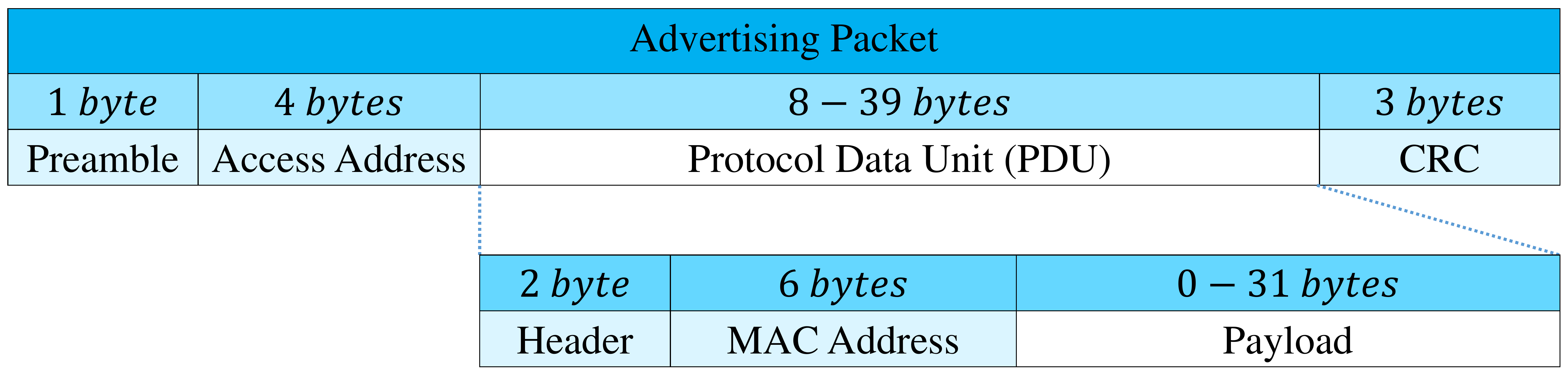}
	\caption{The advertising packet can take up to 47~bytes however, the available bytes for the actual payload is only 31~bytes.}
	\label{fig:advPacket}
\end{figure}  
\subsubsection{Advertising Packet} 
In the non-connectable advertising mode, the smartphone will broadcast the advertising packet over the three advertising channels periodically according to the system-defined advertising interval, $T_a$. The advertising interval defines how frequent a packet is broadcast.  For example, if $T_a = 100~ms$, we shall expect to see at least 10~packets per second. The advertising packet can take up to 47~bytes, as shown in Fig.~\ref{fig:advPacket}. Note that 16~bytes are used for preamble~(1~byte), access address~(4~bytes), header~(2~bytes), MAC address~(6~bytes), and CRC~(3~bytes).
Hence, there are only 31~bytes left to put in the information related to the environmental signature. This poses a question on how to construct a unique yet useful signature that can be encapsulated into this 31~bytes payload.

\subsubsection{Received Signal Strength (RSS)}
\begin{figure}[t!]
	\begin{minipage}{0.48\linewidth}
		\includegraphics[width=\linewidth]{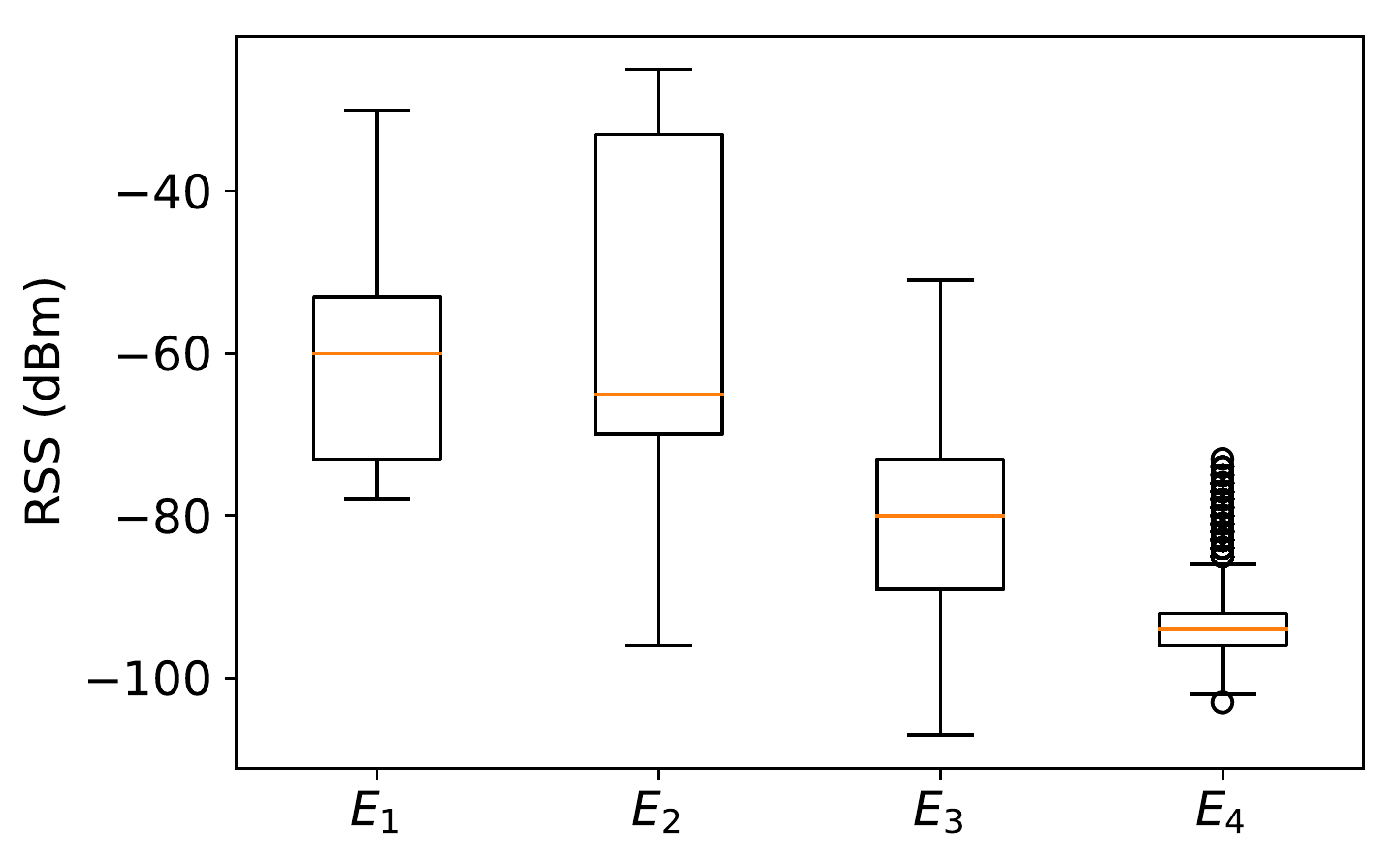}
	\end{minipage}
	\begin{minipage}{0.38\linewidth}
		\begin{footnotesize}
			\begin{tabular}{cl}
				\toprule
				\textbf{\textit{E}} & \textbf{Environment Type}\\
				\midrule
				$E_1$ & Room\\
				$E_2$ & Corridor\\
				$E_3$ & Indoor Open Area\\
				$E_4$ & Outdoor\\
				\bottomrule
			\end{tabular}
		\end{footnotesize}
		\label{table:EnvrType}
	\end{minipage}
	\caption{The variation of RSS values in different environments.}
	\label{fig:envr}
		\vspace{-2ex}

\end{figure}

Following the inverse square law~\cite{fan2012rssi}, the RSS is inversely proportional to the square of the distance. 
Let $P_r$ denotes the signal strength in $dBm$, then:

\begin{equation}
P_r \propto \frac{1}{d^n}
\label{eq:invSqr}
\end{equation}
where $d$ is the distance between any two devices and $n$ is the path loss exponent, and its value is subject to the environmental setting when the measurement is taken. As shown in Fig.~\ref{fig:envr}, different environments have different effects on the RSS variation even though the distance between any two devices in these environments are the same.
Hence, we need to take the environmental factor into consideration when applying the path loss model to estimate the distance given the RSS. 
Section~\ref{sec:proximitySensing} provides a further discussion on our distance estimation approach that addresses the above problem.

\subsection{Smartphone and Embedded Sensors}
The intimacy of smartphones in our everyday life motivates us to adopt the smartphone for contact tracing purposes.
However, there are privacy concerns about using such an intimate device for contact tracing~\cite{kissler2020projecting}.
Many users might worry that their sensitive information which resides in the smartphone will be exposed to the public during the contact tracing.
Our introduced SCT uses the non-connectable advertising mode, hence, none of the neighboring devices are able to connect to the user's device to retrieve any information. Furthermore, we are using a unique environmental signature that contains no information about the user's identity.

Research efforts that tried to address this privacy issue provide better encryption methodologies~\cite{shukla2020privacy},~\cite{bell2020tracesecure}.
However, none of the works discuss the contact tracing in private and public locations. While most users might willingly to participate in contact tracing in the public locations in a hope to flatten the disease spreading curve, they might feel a bit uncomfortable to let the contact tracing application running when they are having their private time in the private location (e.g., home, sleeping room, car, etc.). Future work can be conducted using the embedded sensors on the smartphone to check if the user is in the private or public location. Then, we can use this information to turn on and turn off the contact tracing application accordingly.

%


\begin{figure}[t!]
\centering
\subfloat[Interaction Phase.]
{\includegraphics[width=.7\columnwidth]{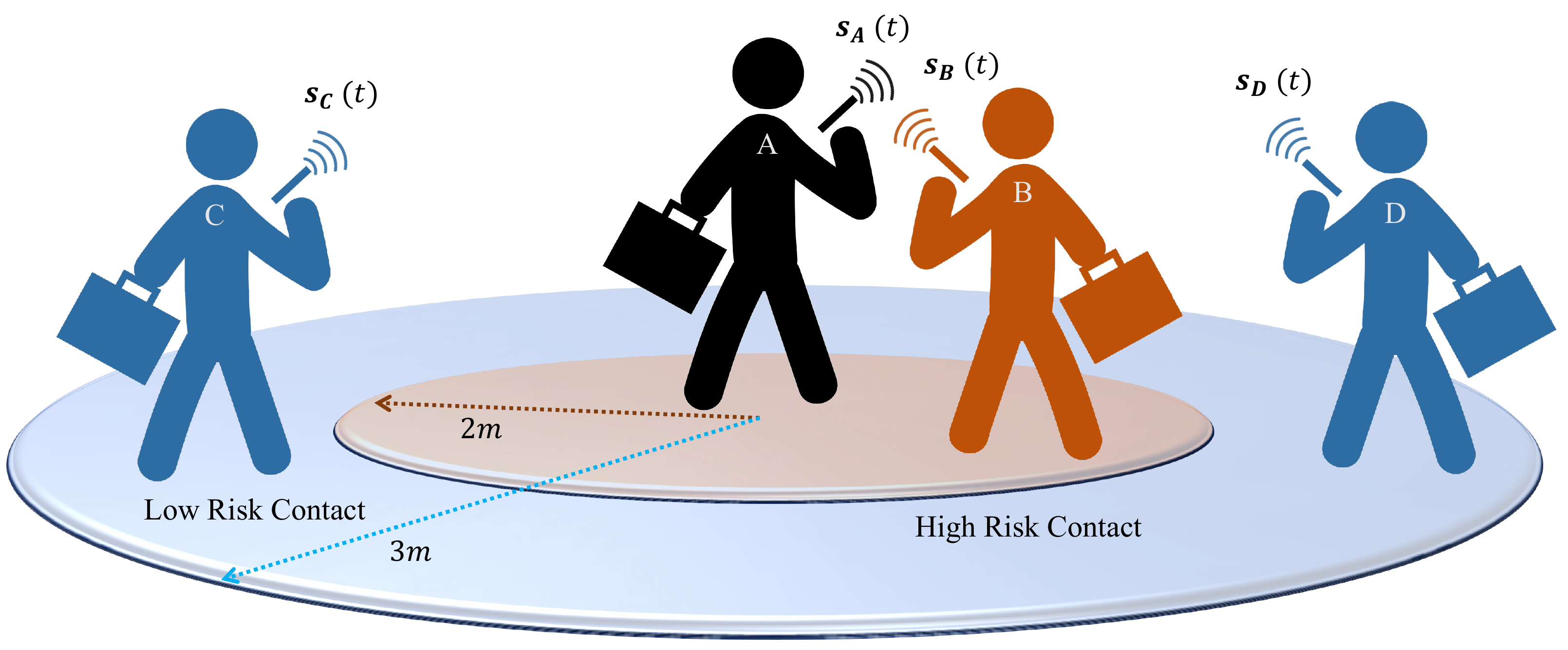}\label{fig:s1_f}}

\subfloat[Tracing Phase.]
{\includegraphics[width=0.9\columnwidth]{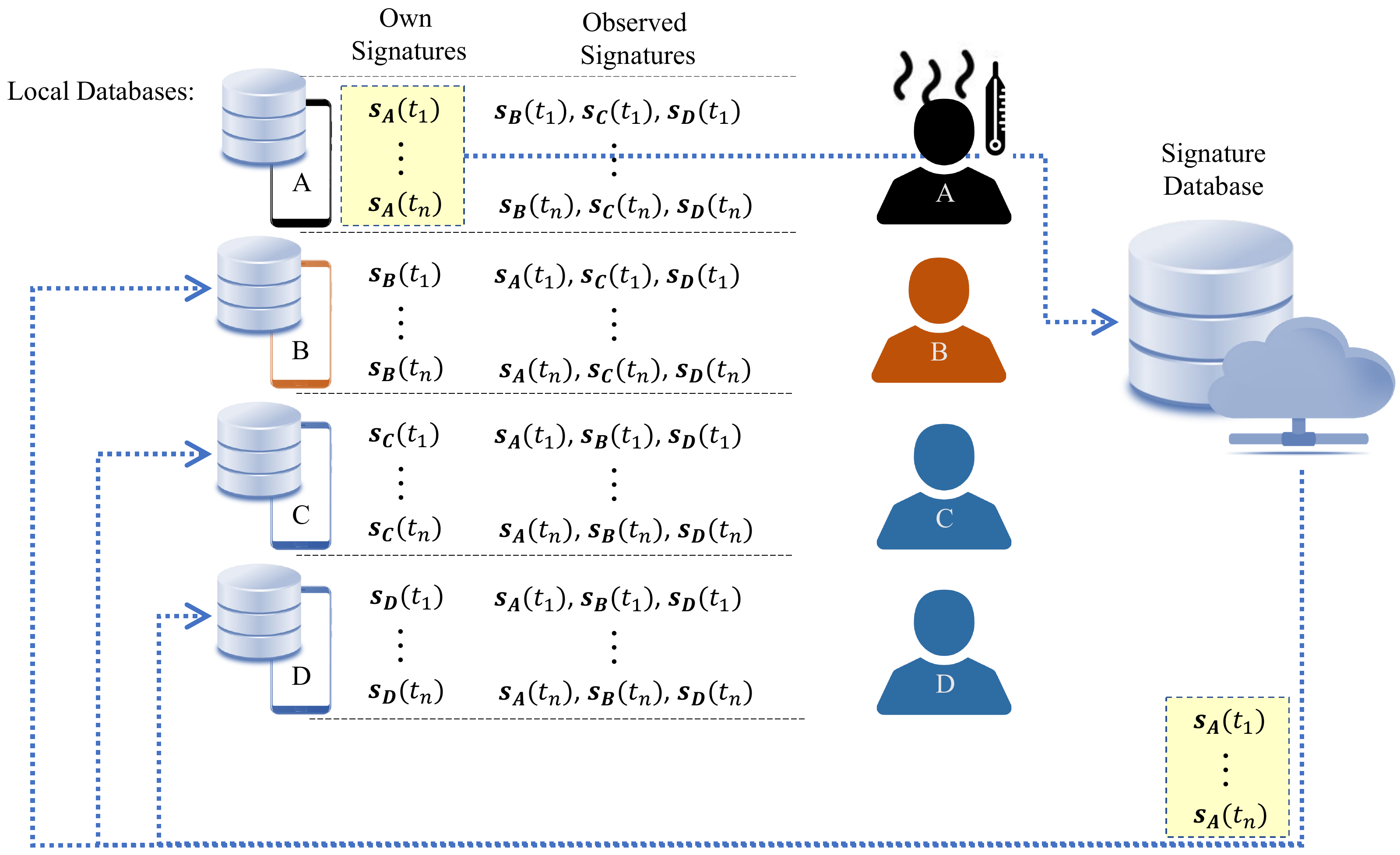}\label{fig:s1_t}}
\caption{Our proposed SCT system consists of 2 phases: (a) interaction phase to log the broadcast and observed signature; and (b) tracing phase to perform the signature matching when a user is diagnosed with the infected disease.}
\label{fig:sctOverall}
	\vspace{-2ex}
\end{figure}

\section{Proposed Smart Contact Tracing System}
\label{sec:sct}
There are two major phases with our SCT system, as shown in Fig.~\ref{fig:sctOverall}: the interaction phase and the tracing phase.
The interaction phase focuses on the following two main components: 1)~privacy-preserving signature protocol, and 2)~precise proximity sensing; whereas the tracing phase aims to provide an efficient signature matching.

\subsection{Interaction Phase}
The interaction phase involves the day to day activities in public locations, such as workplaces, public transports, grocery stores, outdoor parks, etc. 
The contact tracing application starts automatically when it detects the user is in the public location.
The application executes the following functions:
\begin{enumerate}[label=\roman*.]
	\item Signature generation: The smartphone scans for the ambient environmental features. These features are selectively processed to generate a unique signature that can be fit into the 31 bytes advertising payload. The signature will be updated every few minutes.
	\item Signature broadcasting: The smartphone broadcasts the advertising packet containing the unique signature periodically according to the advertising interval of $T_a$. The packet is broadcasted through the non-connectable advertising channels.
	\item Signatures Observation: The smartphone scans the three advertising channels to listen for the advertising packet broadcast by the neighboring smartphones. The scanning is performed in between the broadcasting event.
	\item Proximity sensing: The smartphone measures the RSS values and uses them to  estimate how close it is to the neighboring smartphones. It is assumed to be in proximity when the distance is less than 2~m.
	\item Physical distancing alert: The smartphone triggers a real-time alert to warn the user to keep a healthy distance from nearby users when it detects any physical distancing violation.
\end{enumerate}


\subsection{Tracing Phase}
All the generated signatures and observed signatures will be stored inside the user's local storage, as shown in Fig.~\ref{fig:sctOverall}.
Since the signature does not contain any information about the owner, there is no way for the user to trace or identify the original owner of the observed signatures. 
Furthermore, the signatures are deleted from the local storage permanently once it is expired. We define the expiration period for each signature based on the virus spreading timeframe recommended by the health authorities. For instance, for COVID-19 the expiration period should be 14 days from the day the signature was recorded.  After 14 days, the corresponding signature will be deleted.

If a user is diagnosed with an infectious disease, they can upload all the signatures to the signature database. In Fig.~\ref{fig:sctOverall}, user A uploaded all his signatures to the signature database after he became an infected individual. The database will distribute the signature to all the users' smartphones.  The signature matching computation is taken placed on each individual smartphone and a local alert is triggered when there is a match.
The local alert means that the alert is triggered by the smartphone's program itself, not the centralized alert sent by the server. The server is only used to distribute the data. No program/code is executed on the server to find the close contact. In this way, we can protect the user from revealing their identity and to ensure that none of the match cases can be eavesdropped by malicious hackers. Besides signature matching, the application also performs the classification to classify the potential risk of a user according to the time and distance the user spent with the infected individual.

\section{Privacy-Preserving Signature Protocol}
\label{sec:signature}
While the smartphone can use the non-connectable advertising mode to refuse any incoming connection request attempt, we also need to ensure that the  packet broadcasting will not reveal one's identity.
Several methods have been suggested to protect the user's identity by using an encrypted packet~\cite{PACT, 8422886}. However, these methods require a random generated secret key that can be compromised.  For instance, in TraceTogether application,  a secret key is used to encode the phone number of the user. If the secret key is hacked, then the phone number can be retrieved. In this work, we propose to use the ambient environmental signal to construct a signature vector that can fit into the advertising packet.  

\subsection{Ambient Environmental Features For Signature Generation}
When the application starts the contact tracing, it first generates a signature that can fit into the 31~bytes advertising packet. The signature is a transformed vector containing the ambient environmental features. When the smartphone scans for the packet broadcast by the nearby smartphones, it may also see other BLE devices. For example, in a grocery store, the smartphone might see the BLE beacon attached to the promotional item, the BLE signal from a smartwatch, Apple pencil, smart thermostat, smart lighting control, etc.  The signal strength of each of these devices observed by a user's smartphone is changing depending on the location of the user. Furthermore, some of these devices (i.e., smartwatch, Apple pencil) might not always remain at the same location. Let $P_r(d)$ be a function that returns the time average RSS value measured from a BLE device located at a certain distance from smartphone and $B = \{b_1, b_2, \dots, b_m \}$ be a set of BLE devices excluding the smartphone used for contact tracing, then the observed vector can be expressed as follows:

\begin{equation}
\mathbf{o}_u(t) = (P_e(d_{b_1}), P_e(d_{b_2}), \dots, P_e(d_{b_n}))^T
\end{equation}
where $\mathbf{o}_u(t) \in \mathbb{R}^m$ is an $m$-dimensional vector observed by a smartphone of user $u$ at time $t$. 
The length of the vector $m$ is dependent on the size of $B$. 

Rather than truncating the vector when $m>31$ or filling the vector with some arbitrary values when $m<31$, we define a dictionary $\Psi \in \mathbb{R}^{31 \times m}$ to transform the $m$-dimensional vector to a $31$-dimensional vector.
This dictionary is also known as the secret transformation key to the observed vector.
We can define the dictionary as follows:

\begin{equation}
\Psi =  
\begin{pmatrix}
\psi_{1,1} & \psi_{1,2} & \dots & \psi_{1,m} \\
\psi_{2,1} & \psi_{2,2} & \dots & \psi_{2,m} \\
\vdots     & \vdots     & \dots & \vdots      \\
\psi_{31,1} & \psi_{31,2} & \dots & \psi_{31,m} \\
\end{pmatrix}.
\end{equation}
By multiplying the dictionary with the observed vector, we obtain our unique signature vector.

\noindent 
\begin{definition} \textbf{(Signature Vector)}
	A signature vector $\mathbf{s}(t) \in \mathbb{R}^{31}$ can be obtained by a linear combination of column vectors from  $\Psi$ multiplying with the time average RSS value from each $b_j$ in $B$, i.e.,
	\begin{equation}
	\mathbf{s}(t) = \sum_{b_j \in B} \mathbf{\Psi}_j P_e(d_{b_j}),
	\end{equation}
	where $\mathbf{\Psi}_j = (\psi_{1,j}, \psi_{2,j}, \dots, \psi_{31,j})^T \in \mathbb{R}^{31}$  is the $j$-th column vector from the dictionary. 
\end{definition}

\begin{figure}
	\centering
	\includegraphics[width=0.9\columnwidth]{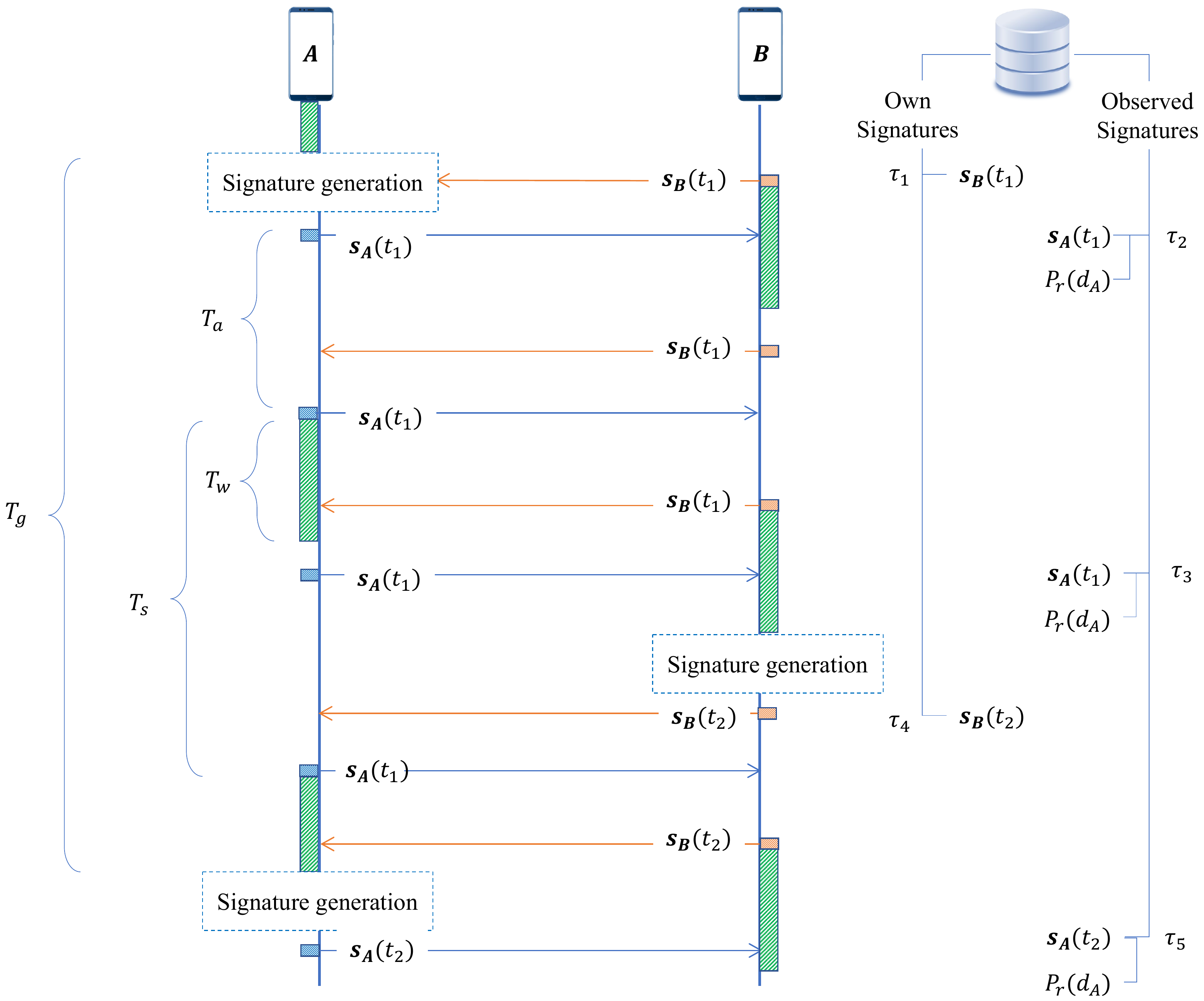}
	\caption{Timing diagram for the advertising, scanning and signature generation activities. All the generated and observed signatures will be logged in the local database, together with a timestamp $\tau$.}
	\label{fig:timeDia}
		\vspace{-2ex}

\end{figure}
 
The dictionary for each smartphone is different. If any two smartphones observe the same ambient features, i.e., similar time average RSS values from a similar set of BLE devices, the generated signature vector is still different.
Note that the above case is very rare because even though both persons appear at the same location at the same time, the RSS values from the same set of BLE devices might be different due to the receiver sensitivity of the smartphone, the smartphone's antenna, the position and orientation of the smartphone, etc.

\subsection{Scanning and Advertising Activities}
Upon the generation of signature, the smartphone encapsulates this signature information into its advertising packet and broadcasts the packet through the non-connectable advertising channels. The advertising interval determines the broadcasting frequency. Suppose that $T_a = 100~ms$, then we should expect about 10 packets per second. However, this also depends on the scanning window and the interval of a smartphone. More precisely, the smartphone can only see the packet when its scanning activity overlaps with the advertising activity. The timing diagram for the advertising, scanning, and signature generation activities is shown in Fig.~\ref{fig:timeDia}. Each activity is triggered periodically according to their interval, i.e., generation interval $T_g$, advertising interval $T_a$, and scanning interval $T_s$.  Given $T_s$, the smartphone will only stay active to listen for the incoming packet for a duration defined by the scanning window $T_w$.

As shown in Fig.~\ref{fig:timeDia}, smartphone A fails to receive $\mathbf{s}_B(t_1)$ from smartphone B, for the first two times since there is no scanning activity in smartphone A when $\mathbf{s}_B(t_1)$ arrived. However, smartphone A manages to receive $\mathbf{s}_B(t_1)$ when smartphone B broadcasts the same packet the third time. According to~\cite{6259791}, the likelihood to see the advertising packet broadcast by neighboring smartphones is high as long as $Ta < T_s$. Intuitively, when the broadcasting frequency is higher than the scanning frequency given the scanning window is sufficiently long, then it is likely for one of the broadcast packets from A meets the scanning windows of B. We can use a continuous scanning (i.e., set $T_w = T_s$) to increase the packet receiving rate. However, such a scanning approach can greatly affect the energy consumption of a smartphone and eventually create an adverse effect on a user's experience. This work mainly focuses on the privacy and preciseness of contact tracing, balancing the energy consumption and packet receiving rate is a possible direction for future work.

The smartphone logs the generated signatures and the observed signatures in its own local storage, as shown in Fig.~\ref{fig:timeDia}.  A timestamp $\tau$ is logged when the smartphone saves a signature into the local database. We could use either a SQL or NoSQL approach to construct this database. The logged timestamp is useful to examine the total time two persons spend in close proximity to each other. Note that for the observed signature, we also log the RSS value upon receiving the packet. This RSS value provides useful information for proximity sensing.

\section{Proximity Sensing and Classification}
\label{sec:proximitySensing}
Proximity sensing has been employed in many scenarios, including identification of the user's proximity to museum collection~\cite{9001059}, and gallery art pieces~\cite{8019467}. There are also works study the proximity detection in dense environment~\cite{8059756}, or proximity accuracy with filtering technique~\cite{8956048}. However, most of these works study the proximity detection between a human and an object attached to a BLE beacon~\cite{8705339}. There is no work studying the proximity sensing between the devices carried by humans.

\subsection{Proximity Sensing with Precise Distance Estimation}
We use RSS to infer the distance between any two smartphones~\cite{liu2014face}. Given the RSS value, the distance of a smartphone to another smartphone that broadcasts the packet can be estimated as follows:

\begin{equation}
d = \exp\left(  \frac{1}{n} \ln(\frac{1}{P_r - c}) \right) 
\label{eq:distRss}
\end{equation}
where $n$ is the path loss exponent and $c$ is the constant coefficient. Both $n$ and $c$ can be obtained through least square fitting. 

Given the distance, then we can determine if the user follows the safe physical distance as recommended by the health authorities. An alert is sent to remind the user if they violate the physical distance rule.

In the distance estimation context, accuracy indicates how close an estimated value to the true value. In other words, the error between the estimated value and true value is close to zero for an accurate estimation. Precision, on the other hand, tells if any two estimated values fall into the same region given similar measurement input (i.e., the RSS value). For contact tracing purposes, an accurate distance estimation is not that critical as compared to precise proximity sensing. We do not need an accurate estimation to tell if the user is in proximity to the infected individual. Rather, a precise estimation is more critical in determining the risk of a user. In particular, we consider that a user belongs to the high-risk group when the user is in close proximity (i.e., $d \leq 2m$) with the infected individual, otherwise, the user is considered to be in the low-risk group.
 

\subsection{Risk Classification}
The problem of classifying the risk of a potential contact can be modeled as a binary hypothesis test.
In particular, consider a risk mapping function $R:(d) \longrightarrow \{ +1, -1 \}$, where $+1$ indicates high-risk and $-1$ low-risk, then there are three hypotheses including the null hypothesis, since we need to consider also the case of false positive and false negative.
False positive is also known as false alarm, in which a user actually belongs to the low-risk group, but the system wrongly classifies them to the high-risk group. False negative, on the other hand, wrongly classifies the user as the low-risk group, but they are actually in close proximity to the infected individual.

Let $H_+$ denote the hypothesis that the user belongs to the high-risk ($+1$) group and $H_-$ the hypothesis that the user belongs to the low-risk ($-1$), then the possible hypotheses are:

\begin{equation}
\begin{aligned}
&H_0: R(d) = 0 \\
&H_+: R(d) = +1  \\
&H_-: R(d) = -1
\end{aligned}
\end{equation}
where $R(d) = 0$ means that the user is not in contact with the infected individual. 
This is valid when the signature matching returns null, which means the user did not encounter any infected individual.

Let $h$, $l$, and $a$ be the ground truth label for high-risk, low-risk and absence (i.e., the user is not in contact with the infected individual, then:
\begin{equation}
\begin{aligned}
&True\ positive: P(H_+ | h) \\
&True\ negative: P(H_0 | a) \\
&False\ positive: P(H_+ | a) + P(H_+ | l) \\
&False\ negative: P(H_- | a) + P(H_- | h) \\
&Miss\ detection: P(H_0 | h) + P(H_0 | l)\\
\end{aligned}
\end{equation}

\begin{figure}[t!]
	\centering
	\includegraphics[width=0.8\columnwidth]{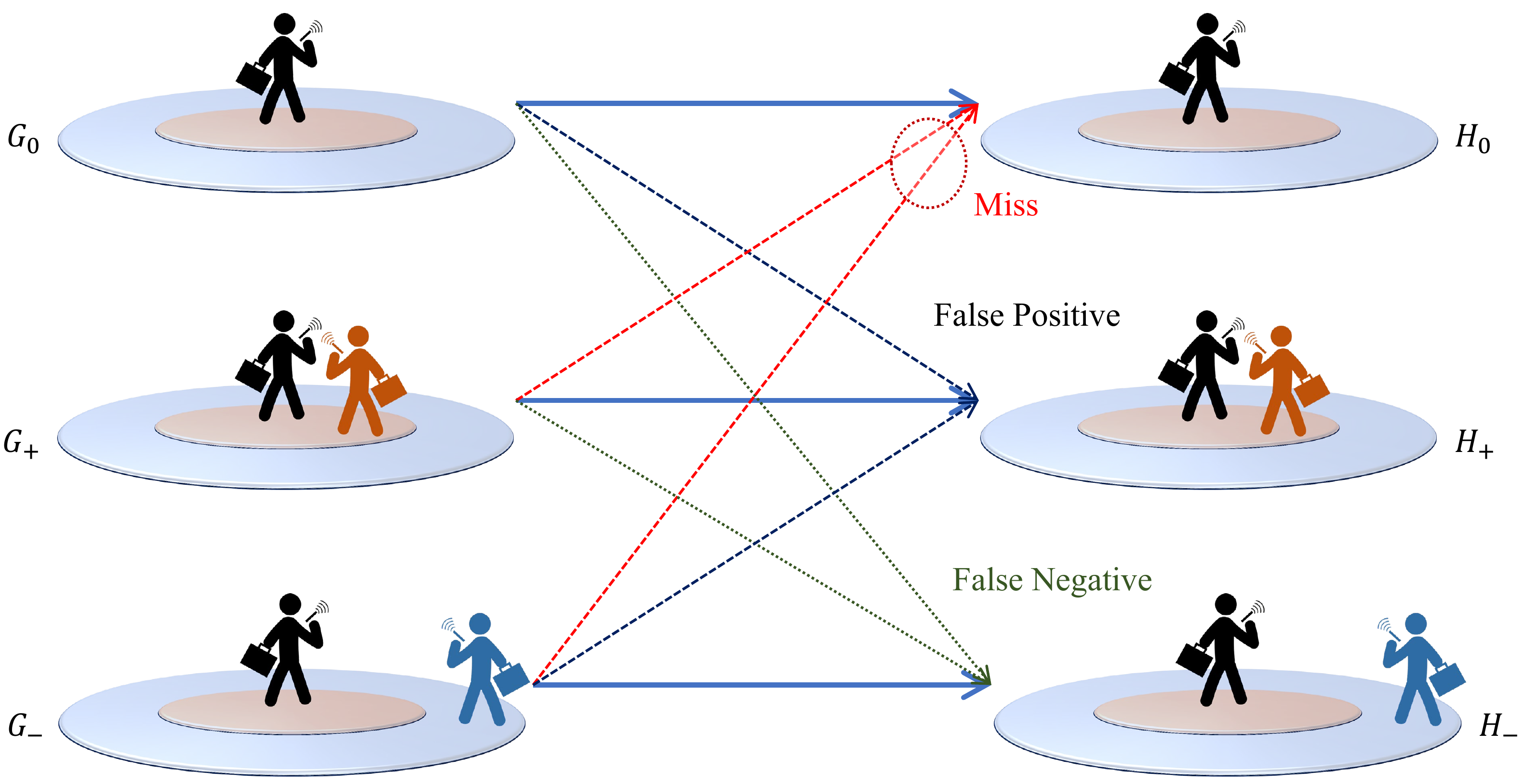}
	\caption{Miss detection,  false positive,  and false negative classification.}
	\label{fig:riskHyp}
\end{figure}

The possible classification outcomes given the above hypotheses are illustrated in Fig.~\ref{fig:riskHyp}. It is obvious that miss detection is undesirable because the user might be at risk but the system considers the user is safe. False negative misclassified the high-risk user to low-risk, but in comparison to miss detection, it can at least detect the user.  However, this may give a wrong impression to the user that the possibility for them to get infected is low, but actually the possibility could be high. False positive  misclassified the low-risk user to high-risk. Even though it is a bit conservative to alarm the user that they are most likely to get infected while they may not, this is a relatively safer outcome than miss detection and false negative.

\section{Implementation and Experiments}
\label{sec:imp}
We have developed a smartphone application to demonstrate our proposed SCT. The application has the following two functions: 1) contact tracing based on the privacy-preserving signature protocol, and 2) physical distancing alert based on precise proximity sensing.  We describe our experimental setup and then discuss our experimental results by comparing the performance with another five classifiers, i.e., decision tree (DT), linear discriminant analysis (LDA), naive bayes (NB), $k$ nearest neighbors (kNN), and support vector machine (SVM).

\subsection{Application Development}
We built an Android Application to demonstrate the functionalities of our proposed SCT. First, the application generates a signature packet according to the privacy-preserving signature protocol. Then, it pushes an alert notification when the user violates the physical distancing rule.  All the generated signatures, observed signatures, and their corresponding signal strengths are all stored in the smartphone's storage. We installed the application into  Android smartphones, including Nokia~8.1 with Android~10, HTC~M9 with Android~7.0 Nougat. At least API~23 is required for the BLE to operate. According to Google, at least 95\% of smartphones support API~23. If a user has a lower API version, then they can still use the application, however, only to receive BLE signals. Other IoT devices, such as BLE beacons can be used from these users to enable them to transmit signals~\cite{iotmag}. When running the application the power requirement is less than 0.24~W which is negligible. For experimental purposes, we created another version of the application that allows us to log the ground truth during the experiment, in order to be able and compare the estimation with the real data.


\subsection{Experimental Setting}
\begin{figure}[t!]
	\centering
	\includegraphics[width=0.7\columnwidth]{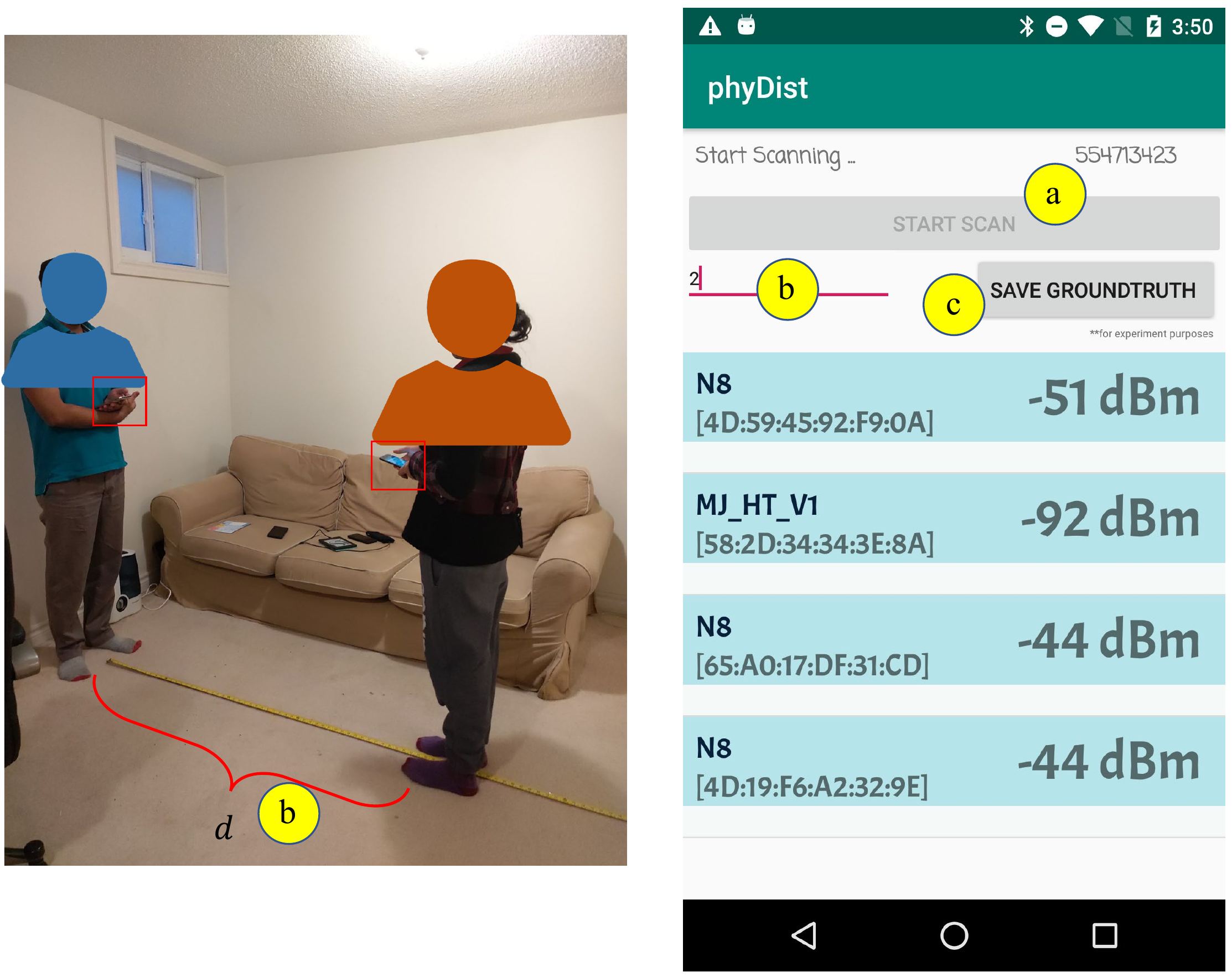}
	\caption{For the experimental purpose, we created another version of the application by (a) adding a manual button to control the start and the end of the experiment, (b) an input field to key in the truth distance measured through the measuring tape, and (c) a save button that save the measurement data.}
	\label{fig:expSetup}
\end{figure}

For experimental purposes,  the application that allows us to log the ground truth was used.
We use the ground truth to evaluate the classification performance.
The following information is logged: the truth distance, name of smartphone, MAC address of BLE chipset, the packet payload, RSS values, time elapsed, and timestamp.
The time elapsed indicates the time difference between the previous broadcast packet and the current broadcast packet, whereas the timestamp is the exact time when the smartphone received the packet.
The true distance is measured with a measuring tape during the experiment, as shown in Fig.~\ref{fig:expSetup}. 

In this experiment, both users are required to hold the smartphone on their hand while doing the measurement.
However, the position of the hand is not fixed, and the user can randomly hold the smartphone according to their own comfort.
The Android BLE API only provides three possible advertising interval settings, as shown in Table~\ref{table:advInt}.
In our experiment, the application is configured to advertise at a ``ADVERTISE\_MODE\_LOW\_LATENCY".
The application is initiated to start the scan when the user presses the scan button. The scan will continue until the user press the button again.
We repeated the experiment for distance from 0.2~m to 2.0~m (with 0.2~m increment each step), and 3~m to 5~m (with 1~m increment each step).
Hence, there are a total of 13 distance points where the measurement is conducted.
For each distance, the application was running for at least 60~s.
All the measurement data are saved into a ``comma-separated values" (.csv) file format and are exported to Matlab for further experiments.

\begin{table}[t!]
	\caption{Advertising interval provided by Android BLE API.}
	\label{table:advInt}
	\centering  
	\begin{tabular}{lx{3cm}rx{3cm}}  
		\toprule
		\cmidrule{1-2}
		Mode     &  $T_a$  \\					
		\midrule
		ADVERTISE\_MODE\_LOW\_LATENCY & $100\,ms$ \\
		ADVERTISE\_MODE\_BALANCED     & $250\,ms$ \\
		ADVERTISE\_MODE\_LOW\_POWER   & $1000\,ms$ \\
		\bottomrule
	\end{tabular}
\end{table}

\subsection{Data Preprocessing}
There are a total of 19,903 data points collected. 
The statistical descriptions of our experimental data are shown in Table~\ref{table:expStats}. We can see that the variance is high at some distance points. This is mostly due to the multipath effects, in which the signal takes multiple paths to reach the receiver. At the receiving end, the signals might be added up constructively or destructively, resulting in many different RSS measurements even the distance is the same. Furthermore, the reflection by moving objects in practical environments might cause undesirable outliers.

\begin{table}[t!]
	\caption{Statistical Descriptions of Experimental Data.}
	\label{table:expStats}
	\centering
	\begin{tabular}{cx{2cm}rx{2cm}rx{2cm}rx{2cm}}
		\toprule
		\cmidrule{1-4}
		Distance ($m$)     &  Count & mean($P_r$)  & var($P_r$)  \\					
		\midrule
		0.2     & 1548     & -58.9994  & 48.8203  \\
		0.4     & 1203     & -62.9967  &  9.8685  \\
		0.6     &  934     & -70.3084  & 10.7666 \\
		0.8     & 1080     & -74.3167  & 16.6930  \\
		1.0     & 1631     & -79.3476  & 14.3153  \\
		1.2     & 1573     & -74.7788  & 12.7322  \\
		1.4     & 3986     & -80.6468  & 41.5620  \\
		1.6     & 1282     & -89.6599  & 11.8577 \\
		1.8     & 1344     & -79.4903  &  4.8413  \\
		2.0     & 1101     & -80.1835  & 15.0263  \\
		3.0     &  886     & -82.1704  & 16.0150 \\
		4.0     & 1220     & -88.5475  & 10.7254  \\
		5.0     & 2115     & -90.4591  & 38.0261  \\
		\bottomrule
	\end{tabular}
\end{table}

\begin{figure}[t!]
	\centering
	\includegraphics[width=0.85\columnwidth]{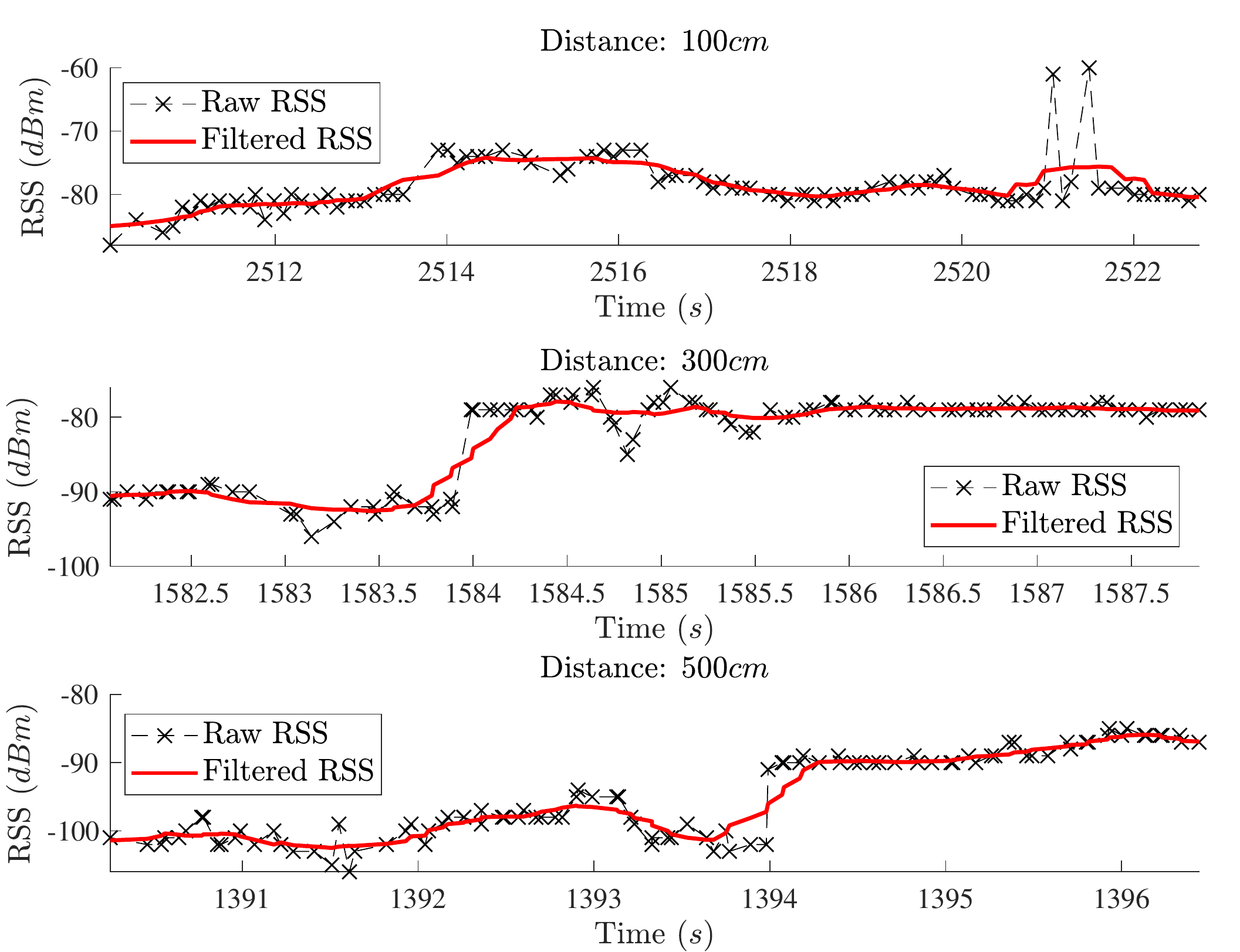}
	\caption{The raw RSS data vs. the filtered RSS data computed through moving average.}
	\label{fig:rawFiltered}
\end{figure}

To mitigate the possible outlier, we apply a moving average filter to the raw data. A comparison between the raw RSS data with the filtered data, for $d = $\{0.2, 1, 2, 3, 5\}~m is shown in Fig.~\ref{fig:rawFiltered}. It is clear that the filtered data provides a much smoother signal.  However, there are still variations in signal strength even though the distance is the same.  In practice, it is hard to obtain only line-of-sight (LOS) signal in indoor environments due to the reflection and diffraction of the signal. Extracting multipath profile features to achieve better distance estimation could be a possible future direction.

\begin{figure*}[t!]
	\centering
	\includegraphics[width=0.8\textwidth]{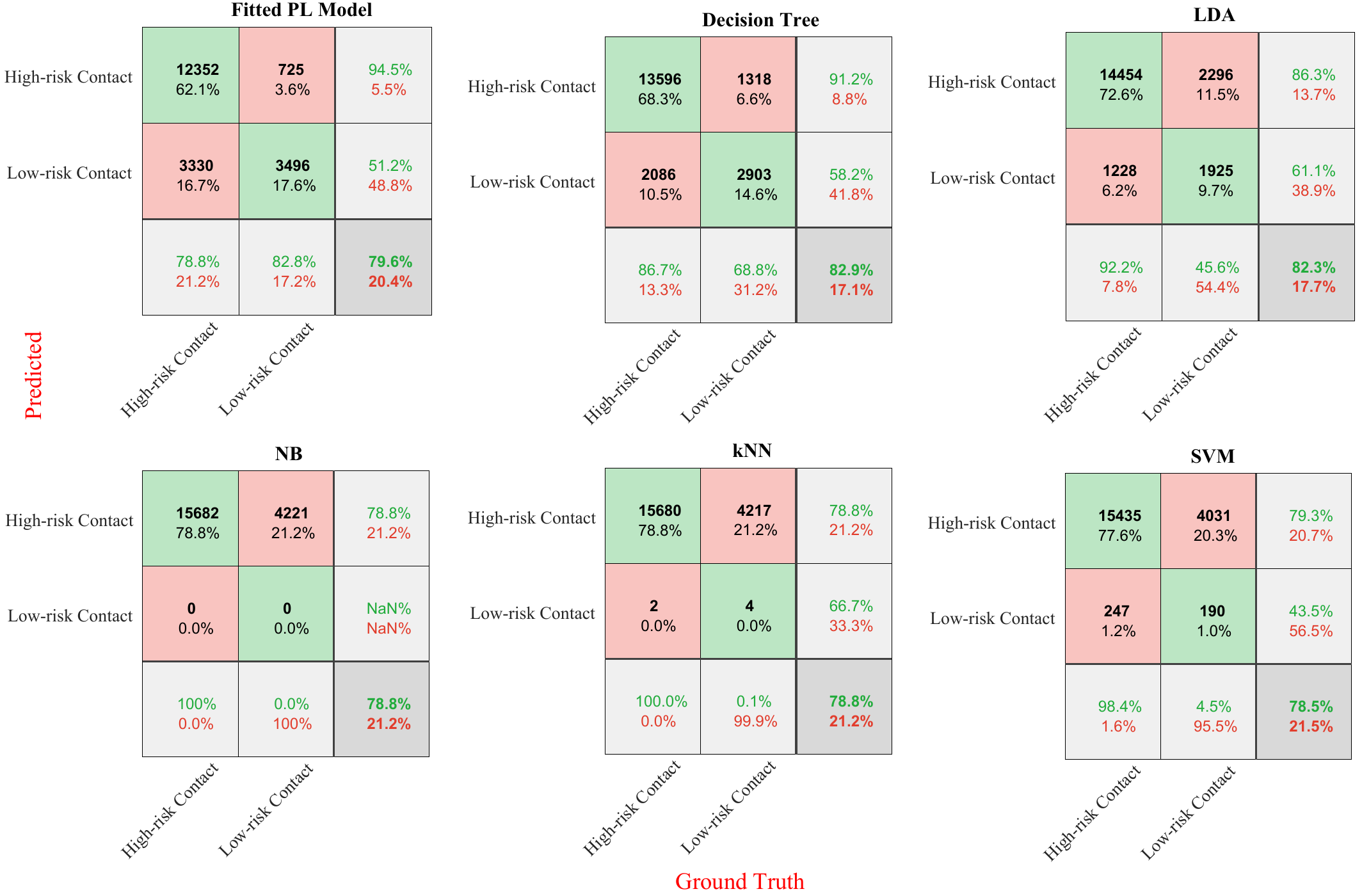}
	\caption{The classification performance of each method is evaluated via a confusion matrix. The x-axis represents the ground truth, whereas the y-axis is the predicted output. The accuracy can be computed by averaging the diagonal elements of the matrix.}
	\label{fig:rawResult}
\end{figure*}
\subsection{Experimental Results}
We applied a non-linear least square fitting to determine the value for coefficients $n$ and $c$, in Eq.~(\ref{eq:distRss}). We used the mean RSS described in Table~\ref{table:expStats} as the dependent variable and distance as the independent variable.
Using this fitted path loss model, we can then estimate the distance by feeding the RSS value measured at each time step into the model.
The ultimate goal is to classify if the user belongs to the high-risk or low-risk group assuming the other user is infected with the disease.
Based on the physical distancing rule recommended by the Canada Health authority~\cite{phyDist},  we classify the user as high-risk if the estimated distance is $\leq 2\,m$ and low-risk if the estimated distance is $> 2\,m$.

We compared the fitted path loss model with five machine learning-based classifiers: DT, LDA, NB, kNN, and SVM.
We separated the measurement data into $80\%$ training data and $20\%$ testing data.
The input RSS was encoded into an 8-bit binary feature.
We used confusion matrix to evaluate the performance of each classifier, as shown in Fig.~\ref{fig:rawResult}.
Overall, DT method achieves the highest accuracy, i.e., $82.9\%$. 
However, if we examined the matrix, we can see that DT also produces a very high false negative rate, i.e., it incorrectly classifies the high-risk group as low-risk for $10.5\%$.
This is not a desirable result for contact tracing purposes because those in the high-risk group are those people that are very likely to get infected but DT method classified them as low-risk.
On the other hand, both NB and kNN have a higher false positive rate, i.e., $21.2\%$.
This is relatively acceptable as it is rather to be more conservative than to be ignored.
While the overall results are acceptable, there is still room for improvement. 
Instead of using the raw measurement, we compared the results with preprocess data. 
Furthermore, we would like to understand how the distancing threshold affects accuracy.

\begin{figure}[t!]
	\centering
	\includegraphics[width=0.6\columnwidth]{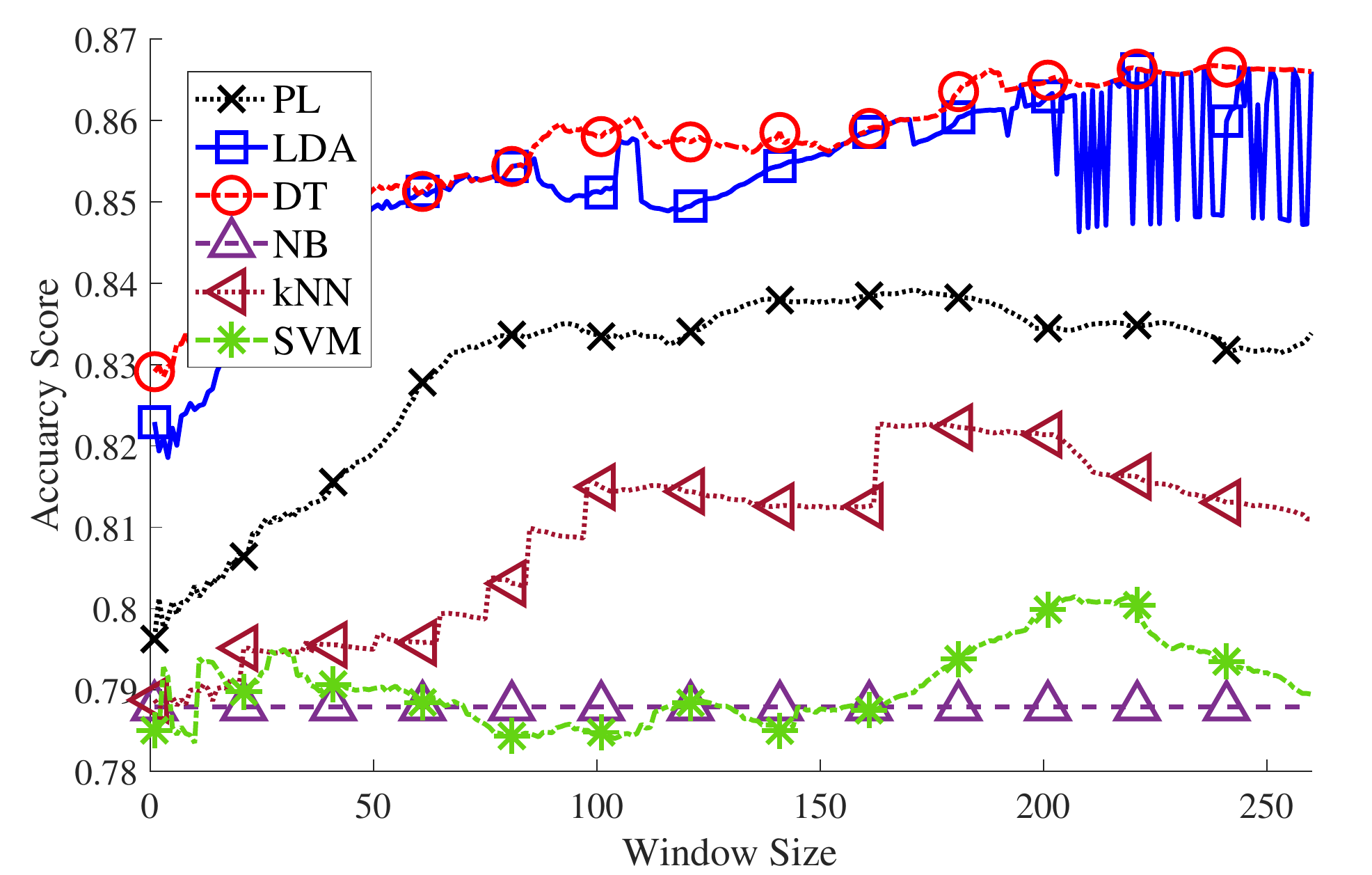}
	\caption{The accuracy score for different window size.}
	\label{fig:accuracy_Samples}
\end{figure}

\begin{figure}[t!]
	\centering
	\includegraphics[width=0.7\columnwidth]{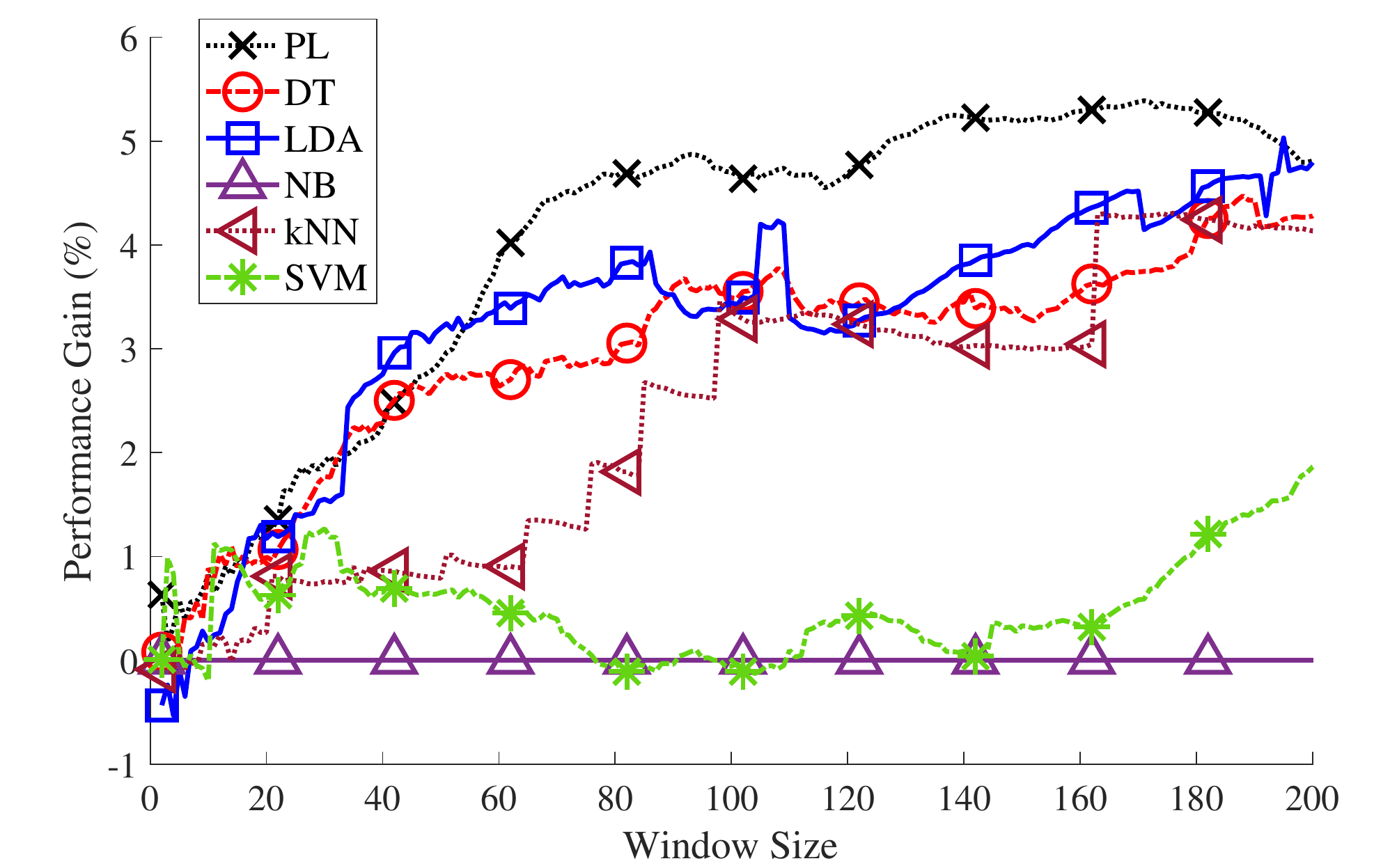} 
	\caption{The performance gain in percentage.}
	\label{fig:window_PerformanceGain}
\end{figure}

\begin{figure*}[t!]
	\centering
	\includegraphics[width=0.7\linewidth]{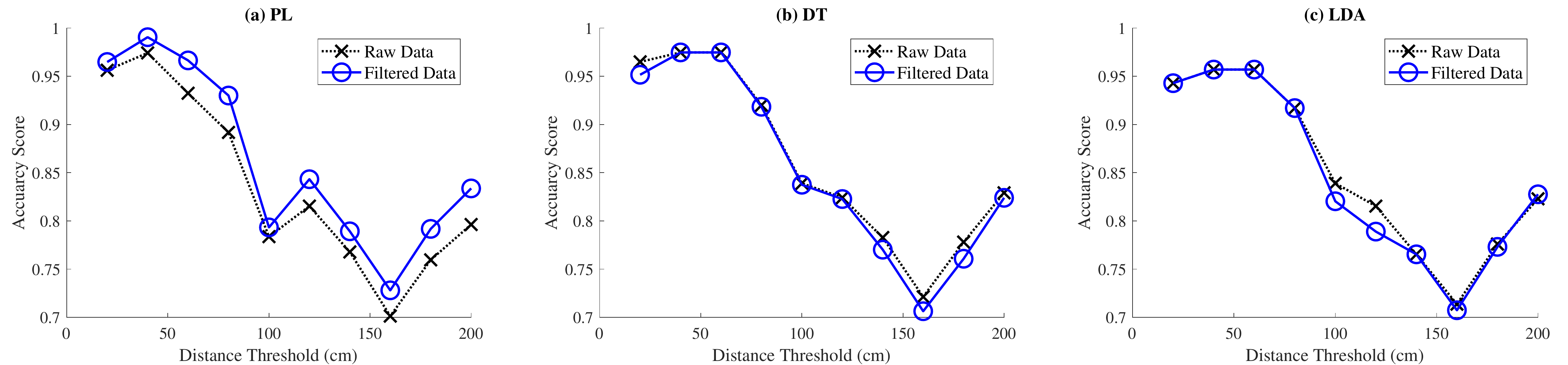}
	\caption{The effect of distance thresholds (i.e., the distance rule used to classify the high-risk and low-risk contact) on the accuracy.}
	\label{fig:distThres}
\end{figure*}

\subsubsection{Implications of Filtered Window}
As shown in Fig.~\ref{fig:rawFiltered}, we can mitigate the possible outliers by preprocessing the data. 
Note that Fig.~\ref{fig:rawFiltered} is obtained by applying the moving average with window size equals to 10.
We further examine the effect of window size on the risk classification performance, and the results are shown in Fig.~\ref{fig:accuracy_Samples}.
It is clear that DT achieves the best performance in comparison to the other approaches when the window size increases.
However, NB does not show any performance gain with increased window size.
LDA, on the other hand, starts to show fluctuation when the window size increases.
This could be due to overfitting during the training process.
Overall, we see that the performance starts to saturate when the window size is more than 100.
The performance gain with respect to window size is shown in Fig.~\ref{fig:window_PerformanceGain}. The performance gain is computed by subtracting the accuracy obtained by the filtered RSS with the accuracy obtained by the raw RSS and then divided by the accuracy obtained by the raw RSS.
From Fig.~\ref{fig:window_PerformanceGain}, we can see that PL is the one that has benefited from the filtered RSS, in which it has higher performance gain than the rest.
In particular, we see that the accuracy obtained via PL is increased from 79.6\% to 83.36\%, which is corresponding to 4.68\% performance gain.
However, not all the methods are benefited from the filtered RSS. Some methods show a performance drop when the window size increases, for example, SVM. Even though DT and LDA can achieve better performance than PL model, both of these methods require extensive training and the accuracy may drop when there are not sufficient data for training.

\subsubsection{Implications of Physical Distancing Threshold}
As discussed previously, our SCT system classifies the contact to high-risk or low-risk according to the physical distancing rule recommended by the health authority.
While we might incorrectly classify the high-risk contact to low-risk due to the fluctuation of the RSS value, we observed that the classification accuracy, in fact, increases when the distancing threshold is smaller.
This is preferable as we would definitely like to classify the user as high-risk when the user is very close to the infected person.
We plotted the accuracy score for different distance thresholds for PL, DT, and LDA. These three methods are selected because they show a good performance, as discussed previously.
We compared the accuracy score between raw data and filtered data, by setting the window size to 100.
This window size is selected based on the window effect on the accuracy score discussed previously.
The accuracy is high when the distancing threshold is less than 1~m, as shown in Fig.~\ref{fig:distThres}.
This result ensures that the system might produce some false negative, but this is mostly happening to the group of users with distance in between 1~m to 2~m from the infected individual.
Interestingly, we also see that the accuracy increases when the distance threshold is increased from 1.5~m to 2~m. 
This indicates that the system is a bit conservative, in which it tends to classify the user as high-risk when the distancing threshold is more than 1.5~m.
Overall, it is safer to have high false positive than high false negative especially if the virus is very contagious.

\subsection{Effect of Smartphone's Positions on the Body} 
\begin{figure}[t!]
	\centering
	\includegraphics[width=0.7\columnwidth]{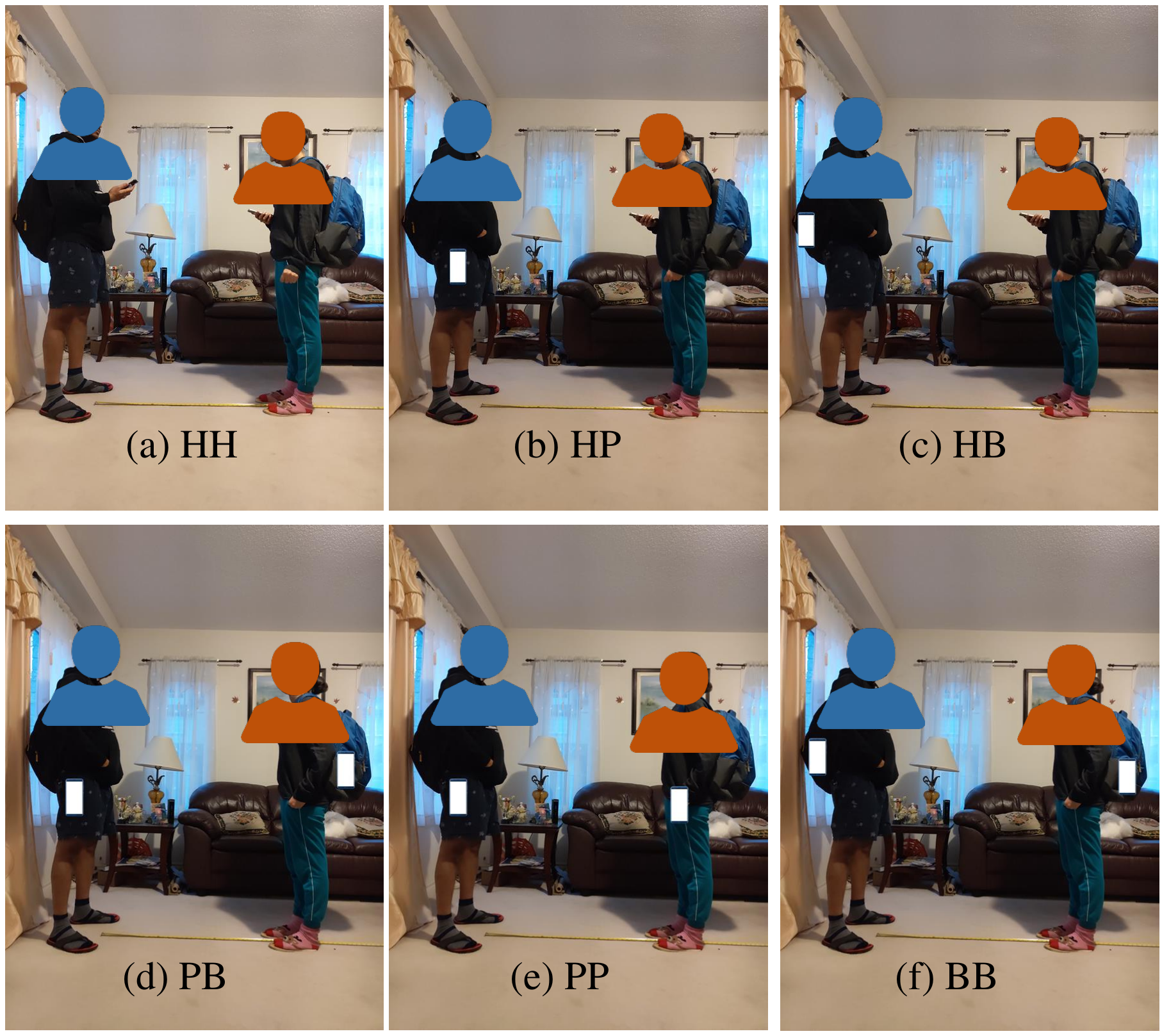}
	\caption{The six combinations of smartphone positions.}
	\label{fig:bodyCombination}
\end{figure}

We extended the experiment to investigate the proximity sensing performance in connection to the positions of smartphone on the body.
The reason is that the user might not carry the smartphone on their hand most of the time.
When they are walking on the street or doing grocery shopping, the user might either carry the phone on their hand, put their phone on their pocket or backpack.
As shown in Fig.~\ref{fig:bodyCombination}, we consider additional five cases on top of the ``Hand-to-Hand (HH)" case discussed previously.
All the measurement data collected from all these six cases can be found in IEEE Dataport~\cite{ieee-dataportRSS} and Github~\cite{githubRSS}. There are a total of 123,718 data points, in which HH contributes 19,903 data points.
The additional five cases and their total data points are listed as follows:
\begin{itemize}
	\item Hand-to-Pocket (HP): one user carries the phone on his/her hand and another user keeps the phone in their pocket. There are a total of 16,081 data points collected for this case.
	\item Hand-to-Backpack (HB): one user carries the phone on his/her hand and another user keeps the phone inside their backpack.
	There are a total of 10,330 data points collected for this case.
	\item Pocket-to-Backpack (PB): one user keeps the phone in the pocket and another user keeps the phone inside their backpack.
	There are a total of 19,161 data points collected for this case.
	\item Pocket-to-Pocket (PP): both users keep their phones in their pockets.
	There are a total of 24,151 data points collected for this case.
	\item Backpack-to-Backpack (BB): both users keep their phones in their backpacks.
	There are a total of 34,092 data points collected for this case.
\end{itemize}

\begin{figure*}[t!]
	\centering
	\includegraphics[width=0.75\linewidth]{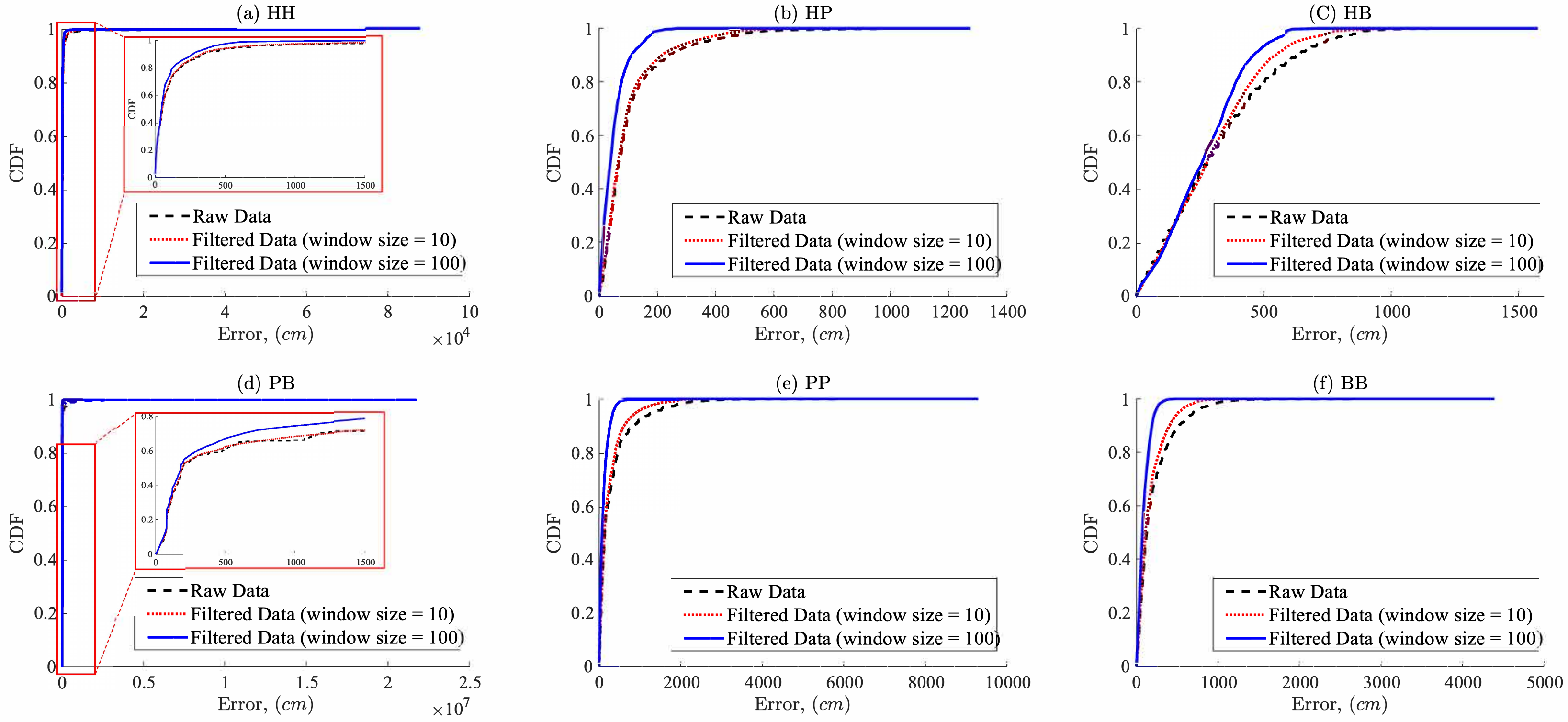}
	\caption{The CDF of the distance estimation errors for all the six cases.}
	\label{fig:all_cdf_png}
\end{figure*}

Previously, we have verified that the accuracy in distance estimation did affect the classification performance.
Hence, we examine the distance estimation performance for all the five cases above. 
In particular, we used the mean absolute error (MAE) to compute the error between the estimated distance and the ground truth distance.

The CDF of the distance estimation errors for all the six cases is plotted, as shown in Fig.~\ref{fig:all_cdf_png}.
It is obvious that the filtered data achieves a better performance. The window size 100 is selected based on the justification provided in Fig.~\ref{fig:accuracy_Samples}.
From the CDF, we can see that, for $80\%$ of the time, the error is less than 1.27~m for HH case, 0.76~m HP case, 3.93~m HB case, 14.94~m PB case, 1.79~m PP case, and 1.48~m BB case.
We observe that PB has the worst performance. This can be explained by the fact that the signals from both smartphones were suffered through different paths of attenuation. 
Hence, even though we tried to calibrate the model based on the environmental factor, the model is unable to capture such variations.
For HH, PP, and BB cases, the signals from both sides might suffer similar path of attenuation.
Take the BB case for example, the smartphone observed a signal blocked by a human body since the smartphone on the other side was located inside the backpack.
Similarly, the smartphone on the other side also observed  similar signal blocked by another human body.
Hence, as long as both smartphones measure the signal from similar position, it is most likely to produce a good estimation.

\begin{table*}[t!]
	\caption{Classification Accuracy for all the six cases}
	\label{table:classificationAccuracy}
	\centering
	\begin{tabular}{c|crc|rc}
		\toprule
		\cmidrule{1-6}
		\multirow{2}{*}{\textbf{Case}}    &  \multirow{2}{*}{\textbf{Method}} & \multicolumn{2}{c|}{\textbf{Raw Data}}   & \multicolumn{2}{c}{\textbf{Filtered Data}}  \\
		        &         & Mean  & 95\% Confidence Interval & Mean  & 95\% Confidence Interval  \\  					
		\midrule
		\multirow{6}{*}{HH} 
		     & DT       &  82.90\% &(82.77\%, 85.64\%)  &  \textbf{85.82}\% &(83.00\%, 85.83\%)  \\
		     & LDA      &  82.12\% &(82.02\%, 85.01\%)  &  85.13\% &(82.23\%, 85.21\%)  \\
		     & NB       &  78.79\% &(78.68\%, 78.79\%)  &  78.89\% &(78.89\%, 78.89\%)  \\
		     & kNN      &  78.81\% &(78.66\%, 81.35\%)  &  81.53\% &(78.96\%, 81.63\%)  \\
		     & SVM      &  78.46\% &(78.45\%, 78.51\%)  &  78.55\% &(78.64\%, 78.66\%)  \\
		     & PL       &  79.62\% &(79.63\%, 83.38\%)  &  83.79\% &(79.63\%, 83.38\%)  \\
		\midrule
		\multirow{6}{*}{HP} 
			& DT       &  82.80\% &(82.69\%, 90.67\%)  &  \textbf{90.75}\% &(82.90\%, 90.84\%)  \\
			& LDA      &  82.17\% &(82.04\%, 90.16\%)  &  90.19\% &(82.30\%, 90.33\%)  \\
			& NB       &  80.27\% &(80.14\%, 81.71\%)  &  81.81\% &(80.40\%, 81.98\%)  \\
			& kNN      &  81.84\% &(81.70\%, 86.95\%)  &  87.15\% &(81.97\%, 87.18\%)  \\
			& SVM      &  18.13\% &(18.00\%, 79.14\%)  &  79.27\% &(18.26\%, 79.39\%)  \\
			& PL       &  77.92\% &(77.92\%, 85.34\%)  &  85.34\% &(77.92\%, 85.34\%)  \\
		\midrule	
		\multirow{6}{*}{HB} 
			& DT       &  77.04\% &(76.87\%, 81.29\%)  &  \textbf{81.44}\% &(77.20\%, 81.60\%)  \\
			& LDA      &  76.99\% &(76.82\%, 78.68\%)  &  78.85\% &(77.16\%, 79.02\%)  \\
			& NB       &  76.70\% &(76.55\%, 76.55\%)  &  76.70\% &(76.86\%, 76.86\%)  \\
			& kNN      &  76.92\% &(76.75\%, 77.82\%)  &  77.99\% &(77.10\%, 78.16\%)  \\
			& SVM      &  23.22\% &(23.06\%, 23.06\%)  &  23.22\% &(23.38\%, 23.38\%)  \\
			& PL       &  23.31\% &(23.31\%, 23.33\%)  &  23.33\% &(23.31\%, 23.33\%)  \\
		\midrule
		\multirow{6}{*}{PB} 
			& DT       &  87.01\% &(87.02\%, 87.45\%)  &  \textbf{87.51}\% &(87.16\%, 87.58\%)  \\
			& LDA      &  87.01\% &(87.02\%, 87.02\%)  &  87.01\% &(87.18\%, 87.18\%)  \\
			& NB       &  87.12\% &(87.05\%, 87.05\%)  &  87.13\% &(87.21\%, 87.21\%)  \\
			& kNN      &  87.03\% &(86.96\%, 86.96\%)  &  87.03\% &(87.10\%, 87.10\%)  \\
			& SVM      &  86.62\% &(86.54\%, 83.45\%)  &  83.53\% &(86.69\%, 83.61\%)  \\
			& PL       &  56.38\% &(56.38\%, 60.37\%)  &  60.37\% &(56.38\%, 60.37\%)  \\
		\midrule	
		\multirow{6}{*}{PP} 
			& DT       &  73.35\% &(73.12\%, 87.18\%)  &  \textbf{87.26}\% &(73.47\%, 87.34\%)  \\
			& LDA      &  72.94\% &(72.82\%, 87.13\%)  &  87.23\% &(73.47\%, 87.34\%)  \\
			& NB       &  71.34\% &(71.22\%, 73.77\%)  &  73.90\% &(73.07\%, 87.33\%)  \\
			& kNN      &  71.41\% &(71.29\%, 73.79\%)  &  73.91\% &(71.47\%, 74.02\%)  \\
			& SVM      &  28.60\% &(28.48\%, 29.54\%)  &  28.60\% &(28.72\%, 29.54\%)  \\
			& PL       &  66.52\% &(66.52\%, 77.34\%)  &  77.34\% &(66.52\%, 77.34\%)  \\
		\midrule	
		\multirow{6}{*}{BB} 
			& DT       &  77.36\% &(77.26\%, 90.78\%)  &  \textbf{90.85}\% &(77.45\%, 90.91\%)  \\
			& LDA      &  76.90\% &(76.81\%, 85.21\%)  &  85.28\% &(76.99\%, 85.36\%)  \\
			& NB       &  76.93\% &(76.85\%, 85.18\%)  &  85.25\% &(77.02\%, 85.32\%)  \\
			& kNN      &  77.16\% &(77.07\%, 78.00\%)  &  78.09\% &(77.26\%, 78.19\%)  \\
			& SVM      &  22.94\% &(22.85\%, 69.34\%)  &  69.44\% &(23.02\%, 69.54\%)  \\
			& PL       &  62.18\% &(62.18\%, 73.28\%)  &  73.28\% &(62.18\%, 73.28\%)  \\
		\bottomrule
	\end{tabular} 
\end{table*}

We also examine the classification performances for all these six cases. 
Table~\ref{table:classificationAccuracy} shows the classification accuracy obtained using the raw data and the filtered data.
From the table, we can see that DT achieves the best performance with more than 81\% accuracy for all the cases.
Compared to the classification based on distance estimation, the machine learning based is more robust to the signal variations caused by body shadowing. Rather than estimating the distance, these classifiers tried to memorize the output given the labeled input during the training process. 
Hence, the amount of data and the validity of data during the training is very important to train a good classifier. 
In general, the machine learning approach can be adopted when there are sufficient training data available, otherwise, the PL model is the best choice for instant proximity sensing.

\begin{figure}[t!]
	\centering
	\includegraphics[width=0.6\columnwidth]{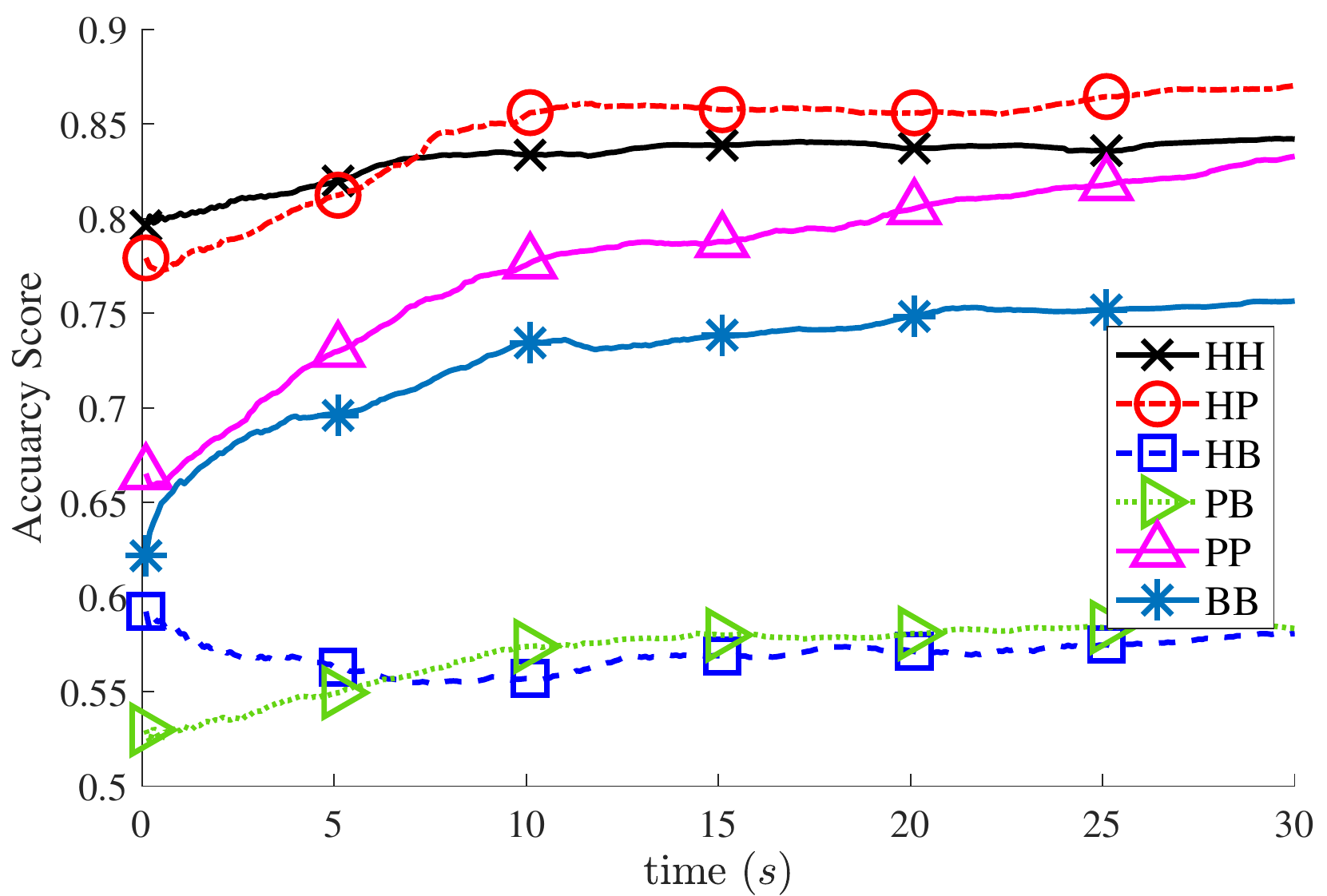}
	\caption{The accuracy obtained over time.}
	\label{fig:all_accTime}
\end{figure}

The accuracy of the PL model increases when the time duration users spent in contact increases, as shown in Fig.~\ref{fig:all_accTime}. 
In general, when the duration increases, the smartphone will be able to observe more signals that help to produce better distance estimation and  increase the classification accuracy.
However, the accuracy starts to saturate after 10~s for most of the cases except HB case.
This result indicates that the smartphone has already observed sufficient RSS data for making the most accurate distance estimation when the time duration is at least  10~s.
The accuracy of the HB case, on the other hand, drops when the time duration increases.
Since the signals arrive at two smartphones from different attenuation paths, the more signals the smartphone observes, the more confusing the smartphone in making a correct estimation.
Note that both HB and PB cases converge to the same accuracy score when the time duration increases.
It is clear that the varying attenuation paths due to the position of smartphones on different body positions can severely affect accuracy.
Future work can be conducted to study the effect of attenuation paths from both sides and come up with an adaptive path loss model that can cater to such diversity in attenuation paths.

\section{Conclusions}
\label{sec:conclusions}
Contact tracing is an essential measure in containing the further spread of a highly infected disease. We propose a smart contact tracing (SCT) system that can provide precise proximity sensing and classify the risk of encountered contact while providing a privacy-preserving signature protocol. From the experimental results, we verified that a BLE- based system for contact tracing is a prominent solution for epidemic control and prevention.  Our SCT system offers tangible results of using RSS values for proximity sensing between two human beings. We have also shared the dataset in an open-source repository to encourage further research.

\bibliographystyle{IEEEtran}
\bibliography{references}

\end{document}